\documentclass{article}

\usepackage{arxiv}

\usepackage[utf8]{inputenc} 
\usepackage[T1]{fontenc}    
\usepackage{hyperref}       
\usepackage{url}            
\usepackage{booktabs}       
\usepackage{amsfonts}       
\usepackage{nicefrac}       
\usepackage{microtype}      
\usepackage{lipsum}		
\usepackage{graphicx}
\usepackage{doi}

\title{A Benchmark to Understand the Role of Knowledge Graphs on Large Language Model's Accuracy for Question Answering on Enterprise SQL Databases}

\author{ Juan F. Sequeda \\
	data.world \\
	\texttt{juan@data.world} \\
 \And
	Dean Allemang \\
	data.world \\
	\texttt{dean.allemang@data.world} \\
  \And
	Bryon Jacob \\
	data.world \\
	\texttt{bryon@data.world} \\
}

\renewcommand{\undertitle}{Technical Report}
\renewcommand{\shorttitle}{Knowledge Graph and LLM Accuracy Benchmark Report}

\hypersetup{
pdftitle={A Benchmark to Understand the Role of Knowledge Graphs on Large Language Model's Accuracy for Question Answering on Enterprise SQL Databases},
pdfauthor={Juan Sequeda, Dean Allemang, Bryon Jacob},
}

\begin{document}
\maketitle

\begin{abstract}

Enterprise applications of Large Language Models (LLMs) hold promise for question answering on enterprise SQL databases. 
However, the extent to which LLMs can accurately respond to enterprise questions in such databases remains unclear, given the absence of suitable Text-to-SQL benchmarks tailored to enterprise settings. 
Additionally, the potential of Knowledge Graphs (KGs) to enhance LLM-based question answering by providing business context is not well understood.
This study aims to evaluate the accuracy of LLM-powered question answering systems in the context of enterprise questions and SQL databases, while also exploring the role of knowledge graphs in improving accuracy. 
To achieve this, we introduce a benchmark comprising an enterprise SQL schema in the insurance domain, a range of enterprise queries encompassing reporting to metrics, and a contextual layer incorporating an ontology and mappings that define a knowledge graph.
Our primary finding reveals that question answering using GPT-4, with zero-shot prompts directly on SQL databases, achieves an accuracy of 16\%. 
Notably, this accuracy increases to 54\% when questions are posed over a Knowledge Graph representation of the enterprise SQL database.
Therefore, investing in Knowledge Graph provides higher accuracy for LLM powered question answering systems.

\end{abstract}


\section{Introduction}
Question answering, the ability to interact with data using natural language questions and obtaining accurate results, has been a long-standing challenge in computer science dating back to the 1960s\cite{10.1145/1460690.1460714,10.1145/800186.810578, 10.1145/355598.362773,10.1145/320251.320253}. 
The field has advanced throughout the past decades \cite{data-atis-original,data-geography-original,data-restaurants-logic}, through Text-to-SQL approaches, as a means of facilitating chatting with the data that is stored in SQL databases\cite{data-sql-imdb-yelp,data-academic,data-atis-geography-scholar,data-restaurants-original,data-restaurants,data-wikisql}.
With the rise of Generative AI and Large Language Models (LLMs), the interest continues to increase. 
These question answering systems hold tremendous potential for transforming the way data-driven decision making is executed within enterprises. 

While question answering systems have shown remarkable performance in several Text-to-SQL benchmarks \cite{finegan-dollak-etal-2018-improving,data-sql-advising}, such as Spider \cite{data-spider}, WikiSQL\cite{data-wikisql}, KaggleDBQA\cite{lee-etal-2021-kaggledbqa} their implications relating to enterprise SQL databases remain relatively obscure. \footnote{A full survey on existing Text-to-SQL benchmarks is outside the scope of this work. We do believe that such survey would be beneficial for the community.}
We argue that existing Question Answering and Text-to-SQL benchmarks, although valuable, are often misaligned with real-world enterprise settings: 
\begin{enumerate}
    \item these benchmarks typically overlook complex database schemas representing enterprise domains, which likely comprise hundreds of tables, 
    \item they also often disregard questions that are crucial for operational and strategic planning in an enterprise, including questions related to business reporting, metrics, and key performance indicators (KPIs), and
    \item a critical missing link is the absence of a business context layer – metadata, mappings, transformations, ontologies, that provides business semantics and knowledge about the enterprise. 
\end{enumerate}

Without these vital components, LLMs for enterprise question answering risk being disconnected from the reality of enterprise data, leading to hallucinations and uncontrolled outcomes. 
Their inability to provide explainable results can significantly impede their trustworthiness and adoption.

Knowledge Graphs (KGs) have been identified as a promising solution to fill the business context gaps in order to reduce hallucinations, thus enhancing the accuracy of LLMs. 
The effective integration of LLMs and KGs has already started gaining traction in academia and industrial research\footnote{https://github.com/RManLuo/Awesome-LLM-KG}\cite{llm_kg}. 
Similarly, from an industry perspective, Gartner states, "Knowledge graphs provide the perfect complement to LLM-based solutions where high thresholds of accuracy and correctness need to be attained."\footnote{Adopt a Data Semantics Approach to Drive Business Value,” Gartner Report by Guido De Simoni, Robert Thanaraj, Henry Cook, July 28, 2023 }.

The goal of this work is to understand the accuracy of LLM-powered question answering systems with respect to enterprise questions, enterprise SQL databases and the role knowledge graphs play to improve the accuracy. 
Specifically, we investigate the extent that LLM can accurately answer Enterprise Natural Language questions over Enterprise SQL databases and to what extent Knowledge Graphs can improve the accuracy.
By assessing the accuracy of question answering systems, we can manage expectations, identify gaps, and focus on areas that require further innovation.

\paragraph{Contribution} In response to these challenges and opportunities, the contributions of our work are the following: 

1) A benchmark consisting of the following:
\begin{itemize}
    \item Enterprise SQL Schema: The OMG Property and Casualty Data Model\footnote{\url{https://www.omg.org/spec/PC/1.0/About-PC}}, an enterprise relational database schema in the insurance domain. 
    \item Enterprise Question-Answer: 43 natural language questions that fall on a combination of two spectrums: 1) low to high question complexity pertaining to business reporting use cases to metrics and Key Performance Indicators (KPIs) questions, and 2) low to high schema complexity which requires a smaller number of tables to  larger number of tables to answer the question. These two spectrums form a quadrant in which questions can be classified: Low Question/Low Schema, HighQuestion/Low Schema, Low Question/High Schema, and High Question/High Schema.
    \item Context Layer: The ontology describing the Business Concepts, Attributes, and Relationships of the insurance domain, and the mappings from the SQL schema to the ontology. The ontology and mappings can be used to create a Knowledge Graph representation of the SQL database.  
\end{itemize}

The benchmark serves as a framework for the results to be reproduced in an enterprise’s own setting using their own enterprise schemas, questions and context. 

2) The results of the benchmark. 
Using GPT-4 and zero-shot prompting, enterprise natural language questions over enterprise SQL databases
achieved 16.7\% accuracy. 
This accuracy increased to 54.2\% when a Knowledge Graph representation of the SQL database was used, thus an accuracy improvement of 37.5\%.
For the questions in each quadrant: 
\begin{itemize}
    \item Low Question/Low Schema, knowledge graph accuracy was 71.1\% while the SQL accuracy was 25.5\%
    \item HighQuestion/Low Schema, knowledge graph accuracy was 66.9\% while the SQL accuracy was 37.4\%
    \item Low Question/High Schema, knowledge graph accuracy was 35.7\% while the SQL accuracy was 0\%
    \item High Question/High Schema, knowledge graph accuracy was 38.7\% while the SQL accuracy was 0\%
\end{itemize}

The results of the benchmark provide evidence that supports our hypothesis that an LLM powered question answering system that answers a Natural Language question over a Knowledge Graph representation of the SQL database returns more accurate results than a LLM 
without a Knowledge Graph. 

The main conclusion of this work is that investing in Knowledge Graphs provides higher accuracy for LLM powered question answering systems.

\section{Research Question and Hypothesis}
\label{sec:ResearchQuestionHypothesis}

The objective of this work is to understand the accuracy of Large Language Models for answering enterprise questions on enterprise SQL databases. 
We investigate the following two research questions: 

\textbf{RQ1:} To what extent Large Language Models (LLMs) can accurately answer enterprise natural language questions over enterprise SQL databases. 

\textbf{RQ2:} To what extent Knowledge Graphs can improve the accuracy of Large Language Models (LLMs) to answer enterprise natural language questions over enterprise SQL databases. 

The hypothesis is the following: 

An LLM powered question answering system that answers a natural language question over a knowledge graph representation of the SQL database returns more accurate results than a LLM powered question answering system that answers a natural language question over the SQL database without a knowledge graph. 

\section{Benchmark Framework}
\label{sec:BenchmarkFramework}

\subsection{Enterprise SQL Schema}
\begin{figure}[hbtp]
\centering
\includegraphics[width=\linewidth]{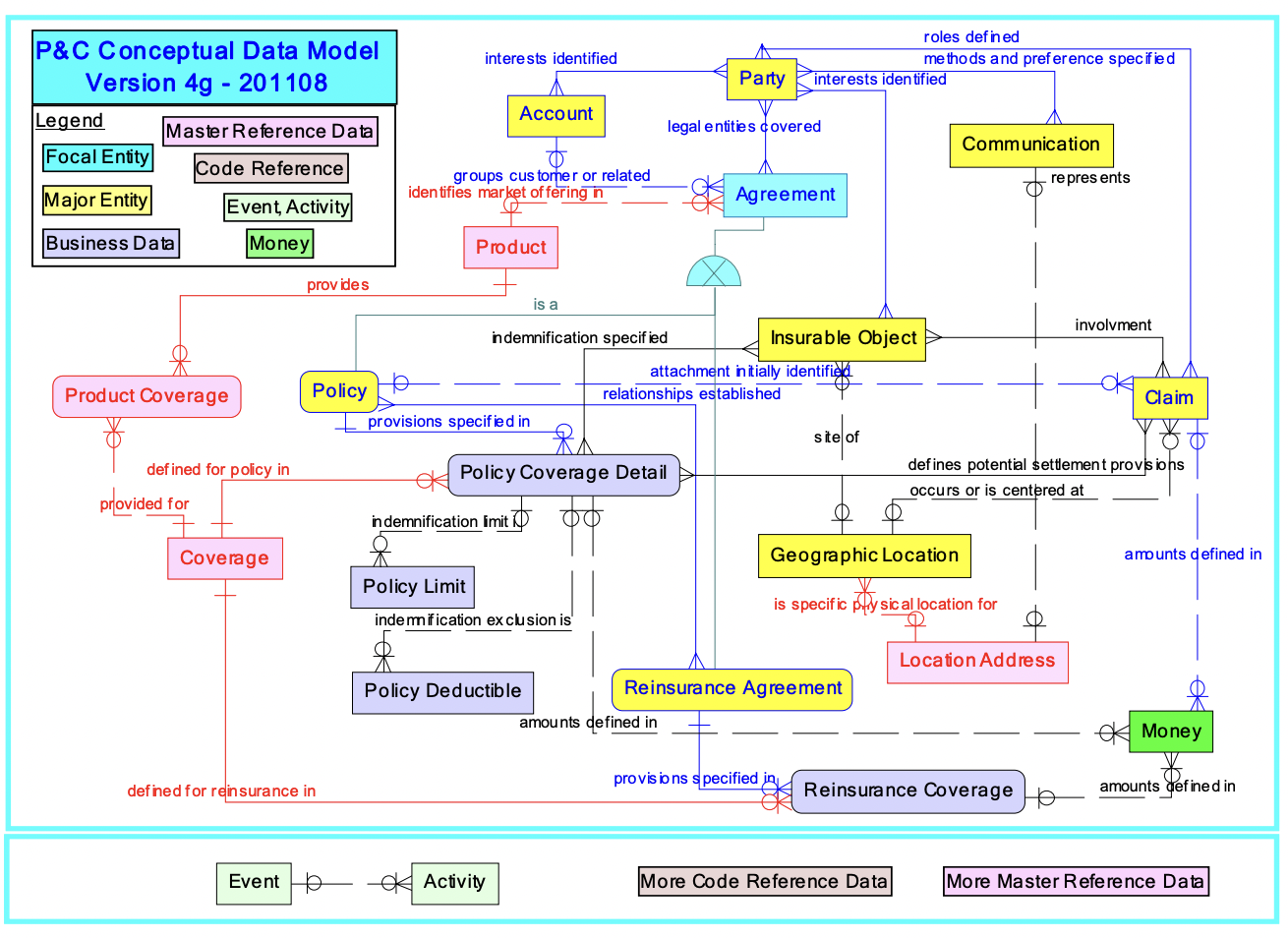}
\caption{P\&C Conceptual Model}
\label{fig:pcmodel}
\end{figure}

The enterprise SQL schema used in the benchmark comes from the P\&C Data Model for Property And Casualty Insurance\footnote{\url{https://www.omg.org/spec/PC/1.0/About-PC}}, a standard model created by Object Management Group (OMG), a standards development organization.  
This OMG specification addresses the data management needs of the Property and Casualty insurance community. The main entities in this Property and Casualty Data Model consist of: Account, Activity, Agreement, Claim, Communication, Coverage, Event, Geographic Location, Insurable Object, Location Address, Money, Party, Policy, Policy Coverage Detail, Policy Deductible, Policy Limit, Product.
Figure \ref{fig:pcmodel} depicts the Property And Casualty conceptual model. 

The physical data model consists of 199 tables. 
Each table has a primary key. 
There are a total of 243 foreign keys. 
The SQL DDL can be found here: \url{https://www.omg.org/cgi-bin/doc?dtc/13-04-15.ddl}

For this current version of the benchmark, we consider a subset of the SQL schema, 13 tables: Claim, Claim\_Amount, Loss\_Payment, Loss\_Reserve, Expense\_Payment, Expense\_Reserve, Claim\_Coverage, Policy\_Coverage\_Detail, Policy, Policy\_Amount, Agreement\_Party\_Role, Premium, Catastrophe.
The schema used in the benchmark can be found in the appendix \ref{Appendix:Schema}. 


\subsection{Enterprise Questions}

The benchmark comes with Question-Answer pairs as evaluation criteria, where the input is the question, and the output is the corresponding answer to the question based on a data instance.
Since there can be multiple valid SQL queries for a given question, the determining accuracy factor is the final output instead of a generated SQL query. 

The questions are classified on a spectrum of low to high complexity:
\begin{itemize}
    \item Low question complexity: Pertains to business reporting use cases, aimed at facilitating daily business operations. From a technical standpoint, these questions are translated into SELECT-FROM SQL queries.
    \item High question complexity: Arises in the context of Metrics and Key Performance Indicators (KPIs) within an organization. These questions are posed to make informed strategic decisions crucial for organizational success. From a technical standpoint, these questions are translated to SQL queries involving aggregations and mathematical functions. 

\end{itemize}

Questions also depend on the number of tables required to provide an answer. Therefore, questions are also classified on a spectrum of low to high schema:
\begin{itemize}
    \item Low schema complexity: Small number of tables (i.e. 0 - 4), denormalized schema
    \item High schema complexity: Larger number of tables (> 4), normalized schema, many-to many join tables, etc.	
\end{itemize}

By combining these two spectrums, four quadrants are defined which are used to classify the questions as shown in the following Figure \ref{fig:question_quadrant}: 

\begin{figure}[hbtp]
\centering
\includegraphics[width=\linewidth]{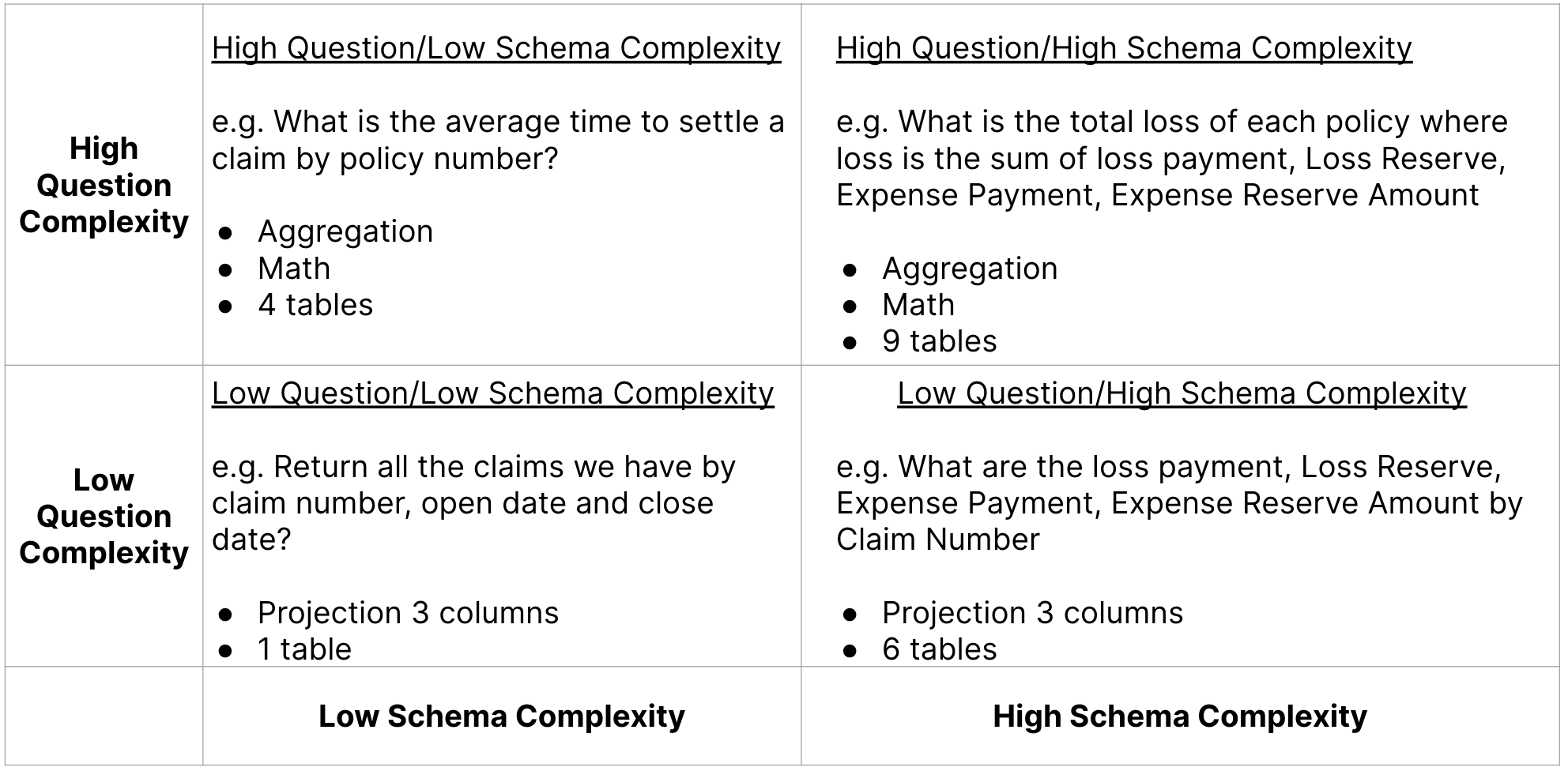}
\caption{Four quadrants to classify questions: (1) Low Question/Low Schema Complexity, (2) High Question/Low Schema Complexity, (3) Low Question/High Schema Complexity, and (4) High Question/High Schema Complexity}
\label{fig:question_quadrant}
\end{figure}

This current version of the benchmark consists of 43 questions. 
The full list of questions can be found in Appendix \ref{Appendix:Questions}

\subsection{Context Layer}

The context layer consists of two parts:
\begin{itemize}
    \item Ontology: Business Concepts, Attributes, and Relationships that describe the insurance domain. 
    \item Mapping: transformation rules from the source SQL schema to the corresponding Business Concepts, Attributes, and Relationships in the target ontology. 
\end{itemize}

For this current version of the benchmark, the context layer is provided in machine readable as RDF: ontology in OWL and mapping in R2RML.
The OWL ontology and R2RML mappings can be used to create the Knowledge Graph either in a virtualized or materialized way. 
The context layer used in the benchmark can be found in the appendix \ref{Appendix:Context}. 
Figure \ref{fig:ontology} is a visual representation of the ontology.

\begin{figure}[hbtp]
\centering
\includegraphics[scale=0.4]{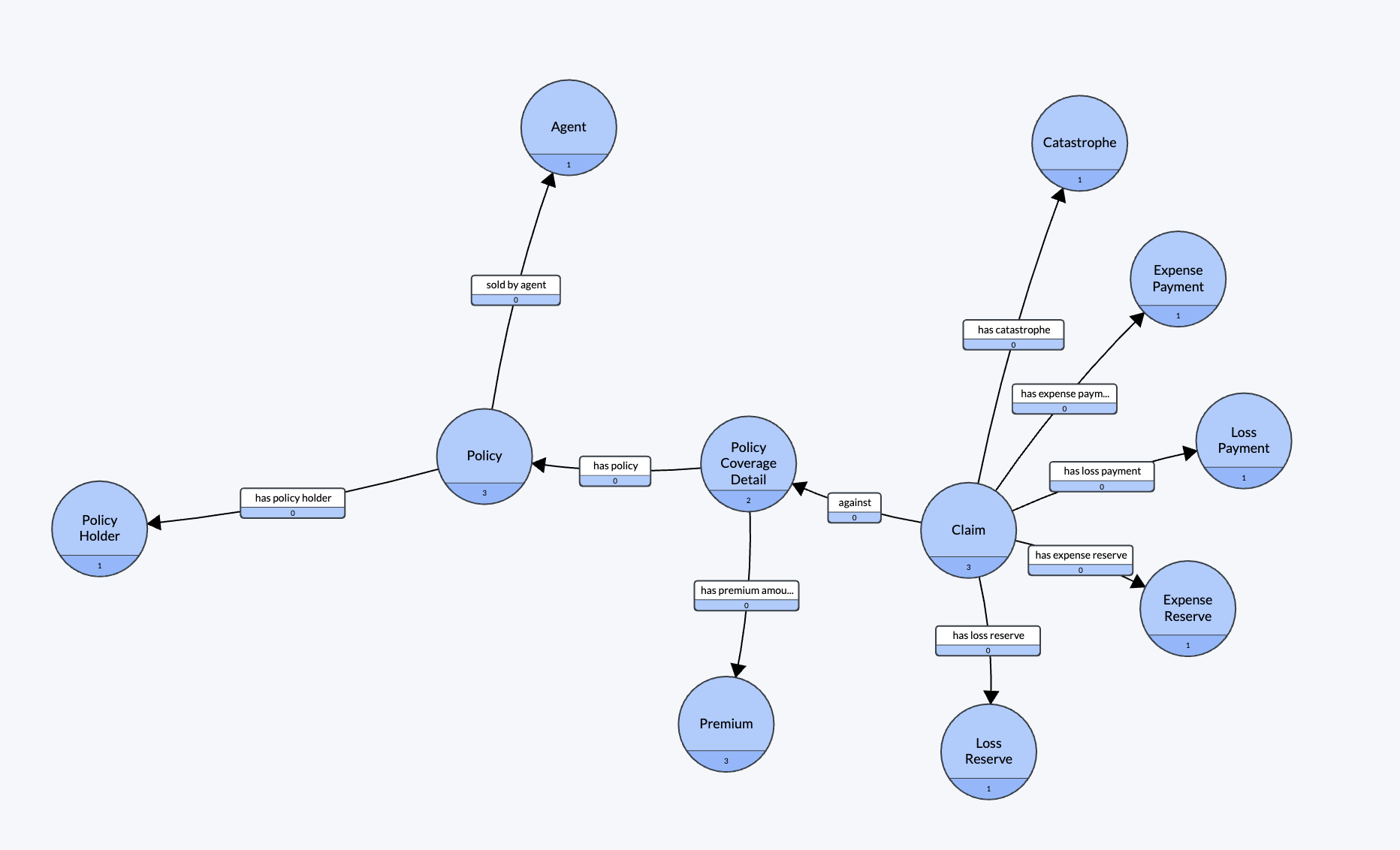}
\caption{Ontology describing relevant parts of the Property and Casualty data model }
\label{fig:ontology}
\end{figure}

\subsection{Scoring}

The benchmark reports three scores: Execution Accuracy, Overall Execution Accuracy and Average Overall Execution Accuracy.

\paragraph{Execution Accuracy (EA)} We follow the metric of Execution Accuracy (EA) from the Spider benchmark \cite{data-spider}. 
An execution is accurate if the result of the query matches the answer for the query. 
Note that the order or the labels of the columns are not taken in account for accuracy.

For example, give the following question-answer pair:  

\textbf{Question:} Return all the claims we have by claim number, open date and close date?

\textbf{Answer:}

\begin{tabular}{|c|c|c|}
\hline
Claim Number & Open Date & Close Date \\
\hline
12312701 & 2019-01-15 & 2019-01-31 \\
\hline
12312702 & 2019-06-02 & 2019-06-27 \\
\hline
\end{tabular}

The following answer is accurate: 

\begin{tabular}{|c|c|c|}
\hline
claim\_number & claim\_close\_date & claim\_open\_date \\
\hline
12312702 & 2019-06-27 & 2019-06-02 \\
\hline
12312701 & 2019-01-31 &  2019-01-15 \\
\hline
\end{tabular}

even though the column headers, the orders of the column and the orders of the rows are different. 

However, the following answer is inaccurate: 

\begin{tabular}{|c|c|c|}
\hline
claim\_id & claim\_close\_date & claim\_open\_date \\
\hline
1 & 2019-01-31 &  2019-01-15 \\
\hline
2 & 2019-06-27 & 2019-06-02 \\
\hline
\end{tabular}

because the answer is missing the claim number, even though it does include claim close data and claim open date. 

\paragraph{Overall Execution Accuracy (OEA)} 
Given the non-deterministic nature of LLMs, there is no guarantee that given an input question, the generated query will always be the same thus providing the same answer. 
Therefore, every question has a Overall Execution Accuracy (OEA) score which is calculated as (\# of EA)/Total Number of runs. 

As an example, based on this scoring: 
\begin{itemize}
    \item 100\% OEA means that every single run generated a query that returned an accurate answer. This means that if there were 100 runs that generated 100 queries, each query returned an accurate result. 
    \item 50\% OEA means that half of the run returned an accurate answer. This means that if there were 100 runs that generated 100 queries, 50 queries returned an accurate result while 50 queries returned an inaccurate result. 
    \item 0\% OEA means that every single run generated a query that returned an inaccurate answer. This means that if there were 100 runs that generated 100 queries, each query returned an inaccurate result. 
\end{itemize}

\paragraph{Average Overall Execution Accuracy (AOEA)}
The Average Overall Execution Accuracy is the average number of OEA scores for a given set of questions. 
This set could be for all the questions in the benchmark or all the questions in a quadrant.

\section{Experimental Setup}
\label{sec:BenchmarkSetup}

The experimental setup is in three parts; the benchmark setup, the question answering system setup and the benchmark processing code.  
This is available on github: \url{https://github.com/datadotworld/cwd-benchmark-data}. 

\subsection{Benchmark setup}

The data setup begins with the P\&C Data Model from the OMG as described above. We have made an excerpt of that model to keep the experiment focused, and to allow the questions to be accessible to people who are not experts in the insurance industry.  We express that DDL as a text file. 

The data itself is expressed  as several tables (CSV format), corresponding to the DDL described above. We have generated sample data for this model, in much the same way as we see in the Spider benchmark\cite{data-spider}.


For the current experiment, we used the 43 questions described in Section \ref{fig:question_quadrant}. 
These questions are written in English, and refer to concepts covered by the data.  
In order to score the execution accuracy of an LLM, we need to have a reference answer to each question. 
An "answer" in this situation is itself a query; it is a query that was written by a human expert, which gives the expected correct answer to the question.  
The query can be in any language that we can run against the data; SQL can be run directly against the data, and SPARQL can be run against the full knowledge graph. 
In this experiment, we provided reference queries in both languages. 
In each case where there is more than one reference query per question, it is imperative that all the reference queries give the same response when run against sample data. 

We want to encourage others to perform experiments using the same data. Toward this end, we have published all of  the benchmark data and metadata as a repository in github. 
The various components in the repository are: 

\begin{itemize}
    \item The DDL for the relational database, as a text file.
    \item The sample data for the tables defined by that DDL.  We publish these in CSV. 
    \item An OWL file (serialized as RDF Turtle) describing the ontology of the knowledge graph.
    \item An R2RML file (serialized as RDF Turtle) describing the mappings from the relational schema to the OWL ontology. 
    \item The questions (natural language) and reference queries (SQL and SPARQL). 
\end{itemize}

Wherever possible, we strive to publish these resources in standard formats.  
The tables themselves are published as CSV, which, while it isn't a standard \textit{per se}, is a format with very broad awareness.  
The data schema is published as a SQL DDL.  
There are many dialects of DDL, but most LLMs are conversant in all of them.  
The knowledge graph context layer components (model and mapping) are published in RDF/TTL, representing OWL and R2RML respectively. 

In the case of the questions and reference queries, we use SQL or SPARQL for the queries, and English for the questions, but we need a way to represent how they fit together, i.e., which reference queries purport to respond to which questions?  
For this purpose, we have built a small ontology of questions and queries that describes how we publish these relationships (in RDF/TTL).  
We also provide a JSON representation of this same information as well. 

For the purposes of the experiments reported in this paper, these materials were hosted on data.world, as follows: 

\begin{itemize}
    \item The DDL is a text file, hosted in data.world as a simple text file.  The data.world API makes this available as is.
    \item The sample data is hosted in data.world as a set of files. data.world treats text files as tables, and provides a SQL interface to query them. 
    \item The OWL file is hosted in data.world as a text file, and is available using the same API as the DDL. Additionally, it is synchronized with the gra.fo visualization tool, which allows it to be manipulated in a form that looks like Figure \ref{fig:ontology}.
    \item The R2RML file is serialized as RDF/TTL, and stored as a file in data.world. By convention, data.world treats any r2rml file as a mapping directive.
    \item The questions and reference queries are hosted in data.world as queries in a workspace. The \texttt{description} field of the query specifies the natural language text of the query. 
\end{itemize}

This setup in data.world provides a number of services that will be helpful as we process the benchmark:

\begin{itemize}
    \item The various text files (the OWL ontology and the DDL) are available through the API directly. 
    \item The sample data stored as multiple CSVs can be queried using SQL directly in the data.world platform (and also through API calls).
    \item The R2RML file is used to drive semantic virtualization; that is, SPARQL queries that refer to concepts in the OWL ontology are translated by the data.world platform into corresponding SQL queries against the data. 
    \item The reference queries can be run by hand, to review their results, or through the API by name.  They can also be downloaded through the API, and the text of the queries can be used in contexts outside data.world (e.g., to run on other database systems). 
\end{itemize}

\subsection{Question Answering System setup}

The question answering system we evaluated was a zero-shot prompt to GPT-4, 
that is instructed to generate a query, which is run against the database.  
The resulting response is compared to the response given by the reference query. 

The particular parameters to the OpenAI API are as follows: 
\begin{itemize}
    \item max\_tokens = 2048
    \item n = 1
    \item temperature = 0.3
\end{itemize}

Additionally, a timeout was set so that computations that take more than 60 seconds are considered to be failures. 

\subsubsection{Question Answering System for SQL}
The question answering system for SQL is shown in Figure \ref{fig:SQlSystem}.  
The question and the SQL DDL for the database are provided as zero-shot prompt to GPT-4. 
These are combined together using the following simple prompt template:

\begin{figure}[hbtp]
\centering
\includegraphics[scale=0.5]{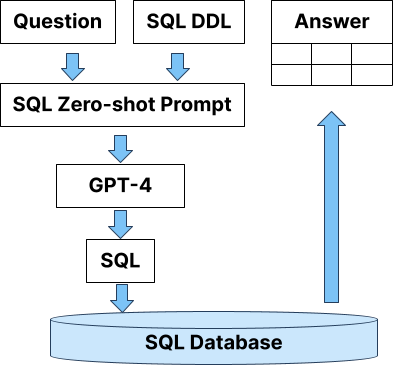}
\caption{Question Answering System for SQL}
\label{fig:SQlSystem}
\end{figure}

\textbf{SQL Zero-shot Prompt}

\fbox{
\begin{minipage}{\linewidth}
Given the database described by the following DDL:

INSERT SQL DDL

Write a SQL query that answers the following question. Do not explain the query.  return just the query, so it can be run verbatim from your response. 

Here's the question: 

"INSERT QUESTION"
\end{minipage}
}

We kept the prompt simple for this experiment, because we wanted to focus on the ability of the contextual information (the DDL in the case of SQL) to provide necessary information for the formation of the query. 

The resulting query is sent verbatim to the SQL processor of data.world, which returns an answer in a tabular form.  
This is converted into a Pandas DataFrame for comparison. 
At the same time, the reference query for the question is sent to data.world, and its result is also converted to a DataFrame.  
Once they are both in the form of DataFrames, it is a simple matter to compare them.  
Details of this comparison are available from the Spider project\cite{data-spider}. 

\subsubsection{Question Answering System for Knowledge Graph}

The question answering system for the Knowledge Graph is shown in Figure \ref{fig:SPARQLSystem}.  
The question and the OWL ontology are provided as zero-shot prompt to GPT-4. 
These are combined together using the following simple prompt template:

\begin{figure}[hbtp]
\centering
\includegraphics[scale=0.5]{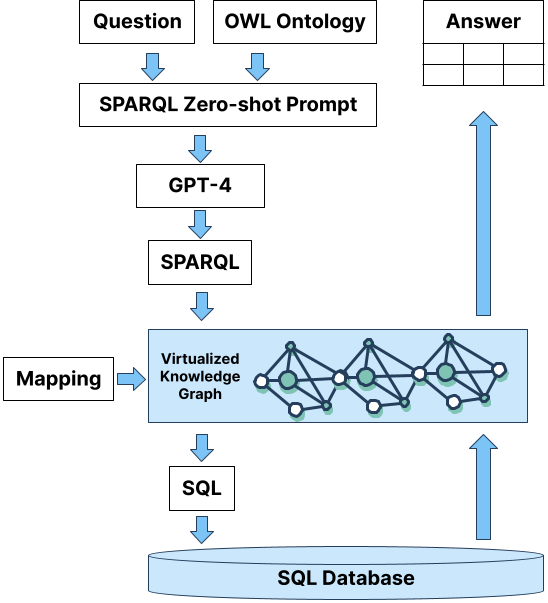}
\caption{Question Answering System for Knowledge Graph}
\label{fig:SPARQLSystem}
\end{figure}

\textbf{SPARQL Zero-shot Prompt}

\fbox{
\begin{minipage}{\linewidth}
Given the OWL model described in the following TTL file:

INSERT OWL ontology

Write a SPARQL query that answers the question.  The data for your query is available in a SERVICE identified by <{mapped}>.  Do not explain the query.  return just the query, so it can be run verbatim from your response.

Here's the question: 

"INSERT QUESTION"
\end{minipage}
}

As in the SQL case, we kept the prompt simple.  
The extra line about the SERVICE allows the LLM to produce queries that invoke the data.world knowledge graph virtualization layer.  
In principle, this adds some complexity to the SPARQL prompt, but in practice, GPT-4 seemed to handle it very well. 

The resulting query is sent verbatim to the SPARQL processor of data.world, and the result converted to a DataFrame, just as for the SQL case.

\subsection{Processing setup}

The point of the experimental setup is to gather the data needed to compute execution accuracy (from which we can compute the derivative metrics of Overall Execution Accuracy and Average Overall Execution Accuracy). 
The basis of Execution Accuracy is to ask the Question Answering system to generate a query, execute that query, and compare the results of the generate query to the results given by the corresponding reference query.  
In principle, there could be a number of outcomes for this comparison. 
The two most interesting are when the results match (a successful execution), and when they do not match (an unsuccessful execution).  
Other cases involve various errors; the generated query could result in a syntactically incorrect query; the Question Answering system could fail to provide a query at all (due to timeout or network failure), and even pathological situations in which the reference query fails for some systemic reason. 
In this setup, we treat all of the failure modes as unsuccessful execution, so the comparison is binary; either it succeeds or it fails. 

The results are recorded as a sequence of records called "Runs"; each of these is represented in RDF.  A run includes a reference to the original question, the query produced by the generator, the language requested (for this experiment, that's SQL or SPARQL),  a timestamp for when the run was completed, and a boolean status of True (for successful matches) and False (for unsuccessful matches). 

The data used in this report was collected between September 20, 2023 and October 13, 2023.  The LLM can take some time to run, and with 43 questions, one Run in SQL and one in SPARQL, an entire experimental session involves 86 calls to the LLM, and 172 query executions.  For this reason, we ran sessions on a cron schedule at various down times for computing resources over the course of a few weeks.

\section{Results}
\label{sec:Results}

The results are presented in four parts 1) overall, 2) question quadrant, 3) partial accuracy and 4) inaccurate results. 
In the results, we refer to
\begin{itemize}
    \item SPARQL as question over Knowledge Graph representation of the SQL database and 
    \item SQL as questions directly on the SQL databases without a Knowledge Graph.
\end{itemize}

Given that the OEA of a question is a percentage, the results are presented as a heatmap. 
Every cell corresponds to a generated query for the given question. The value in the cell is the OEA for that question. 
The green color corresponds to 100\% OEA. The red color corresponds to 0\% OEA. The color scale goes from green to red.

The Overall and Quadrant results are presented in Table\ref{table:Results}.

\renewcommand{\arraystretch}{1.2}
\begin{table}[h!]
\centering
\begin{tabular}{|l|c|c|c|}
\hline
\ & \textbf{w/o KG (SQL)} & \textbf{w/ KG (SPARQL)} & \textbf{Improvement}\\
\hline
\textbf{All Questions} & 16.7\% & 54.2\% & 37.5\%\\
\hline
\textbf{Low Question/Low Schema} & 25.5\% & 71.1\% & 45.6\%\\
\hline
\textbf{High Question/Low Schema} & 37.4\% & 66.9\% & 29.5\%\\ 
\hline
\textbf{Low Question/High Schema} & 0\% & 35.7\% & 35.7\%\\
\hline
\textbf{High Question/High Schema} & 0\% & 38.5\% & 38.5\%\\
\hline
\end{tabular}
\caption{Average Overall Execution Accuracy (AOEA) of Overall and Quadrant Results}
\label{table:Results}
\end{table}

\subsection{Overall}
\begin{figure}[hbtp]
\centering
\includegraphics[scale=0.3]{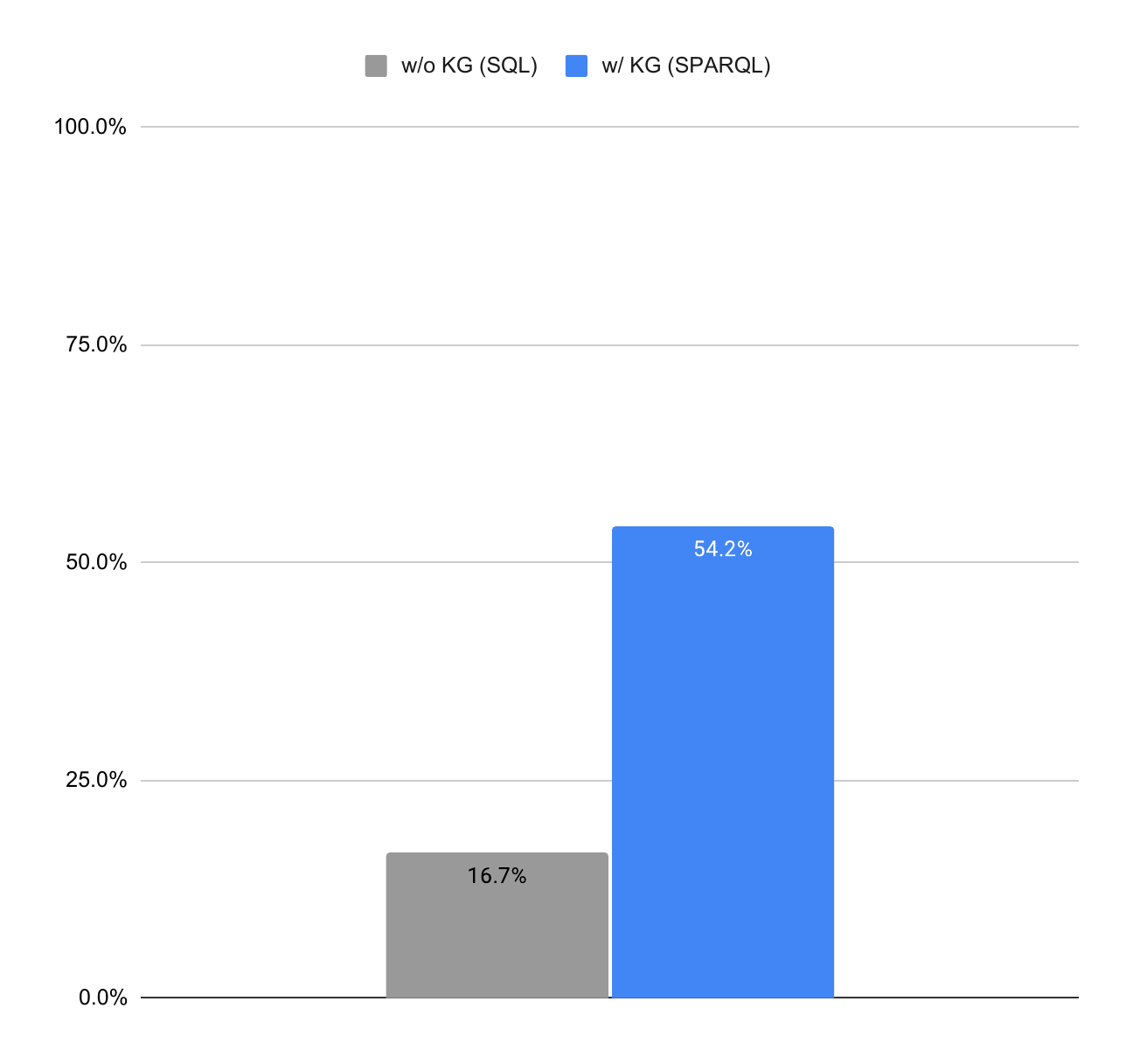}
\caption{Average Overall Execution Accuracy (AOEA) of SPARQL and SQL for all the questions in the benchmark}
\label{fig:fig1}
\end{figure}

By grouping all the questions in the benchmark, SQL achieves an AOEA of 16.7\%. 
In comparison, SPARQL achieves an average OEA of 54.2\% as shown in Figure \ref{fig:fig1}. 
Therefore, overall SPARQL accuracy was 3x the SQL accuracy.

The heatmap that depicts the OEA for each question is shown in the following figure \ref{fig:fig2}:

\begin{figure}[hbtp]
\centering
\includegraphics[scale=0.5]{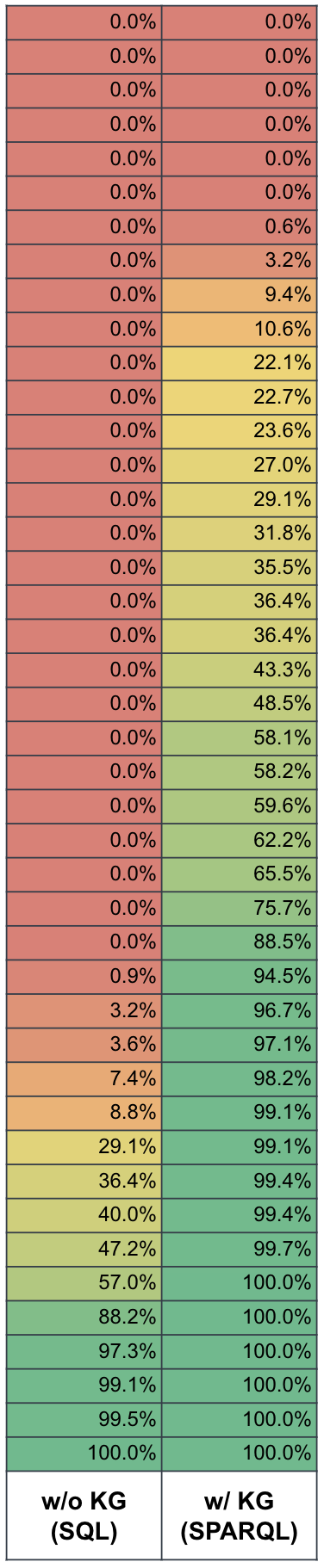}
\caption{Overall Execution Accuracy (OEA) of SPARQL and SQL for all the questions in the benchmark as heatmap}
\label{fig:fig2}
\end{figure}

\paragraph{Discussion}
Overall, Natural Language questions translated to SPARQL over a Knowledge Graph representation of the SQL database achieved 3x the accuracy of natural language questions translated to SQL and executed directly over the SQL database. 
Thus, this result is evidence that supports our hypothesis: 
\textit{An LLM powered question answering system that answers a Natural Language question over a Knowledge Graph representation of the SQL database returns more accurate results than a LLM powered question answering system that answers a Natural Language question over the SQL database without a knowledge graph.}

Recall that the goal is to understand \textit{to what extent} an LLM is accurate. 
These results provides a first understanding of that extent.
We acknowledge that combining all the questions into one overall result is not satisfactory because there are nuances to the types of questions. 
This is why we also present the results in each of the quadrants.

\subsection{Quadrant}

Figure \ref{fig:fig3} presents the AOEA scores for questions in each quadrant as a:

\begin{figure}[hbtp]
\centering
\includegraphics[scale=0.35]{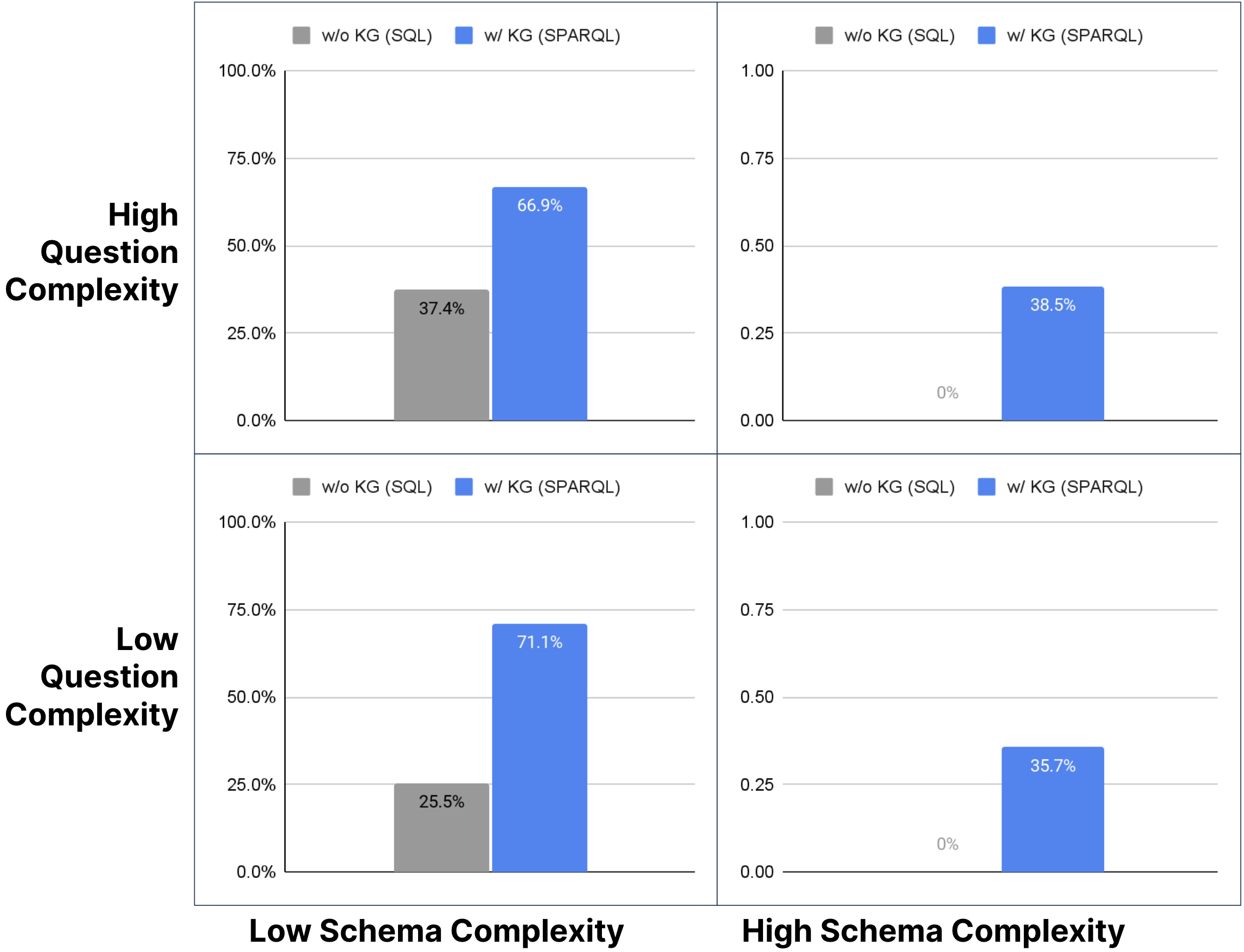}
\caption{Average Overall Execution Accuracy (AOEA) of SPARQL and SQL for each quadrant}
\label{fig:fig3}
\end{figure}

We observe the following results: 
\begin{itemize}
    \item Low Question/Low Schema: SQL achieves an AOEA of 25.5\%. In comparison, SPARQL achieves an AOEA of 71.1\%. The SPARQL accuracy is 2.8X the SQL accuracy.

    \item High Question/Low Schema: SQL achieves an AOEA of 37.4\%. In comparison, SPARQL achieves an AOEA of 66.9\%. The SPARQL accuracy is 1.8X the SQL accuracy.

    \item Low Question/High Schema: SQL was not able to answer any question accurately. In comparison, SPARQL achieves an AOEA of 35.7\%.

    \item High Question/High Schema: SQL was not able to answer any question accurately. In comparison, SPARQL achieves an AOEA of 38.7\%.

\end{itemize}

\begin{figure}[hbtp]
\centering
\includegraphics[scale=0.35]{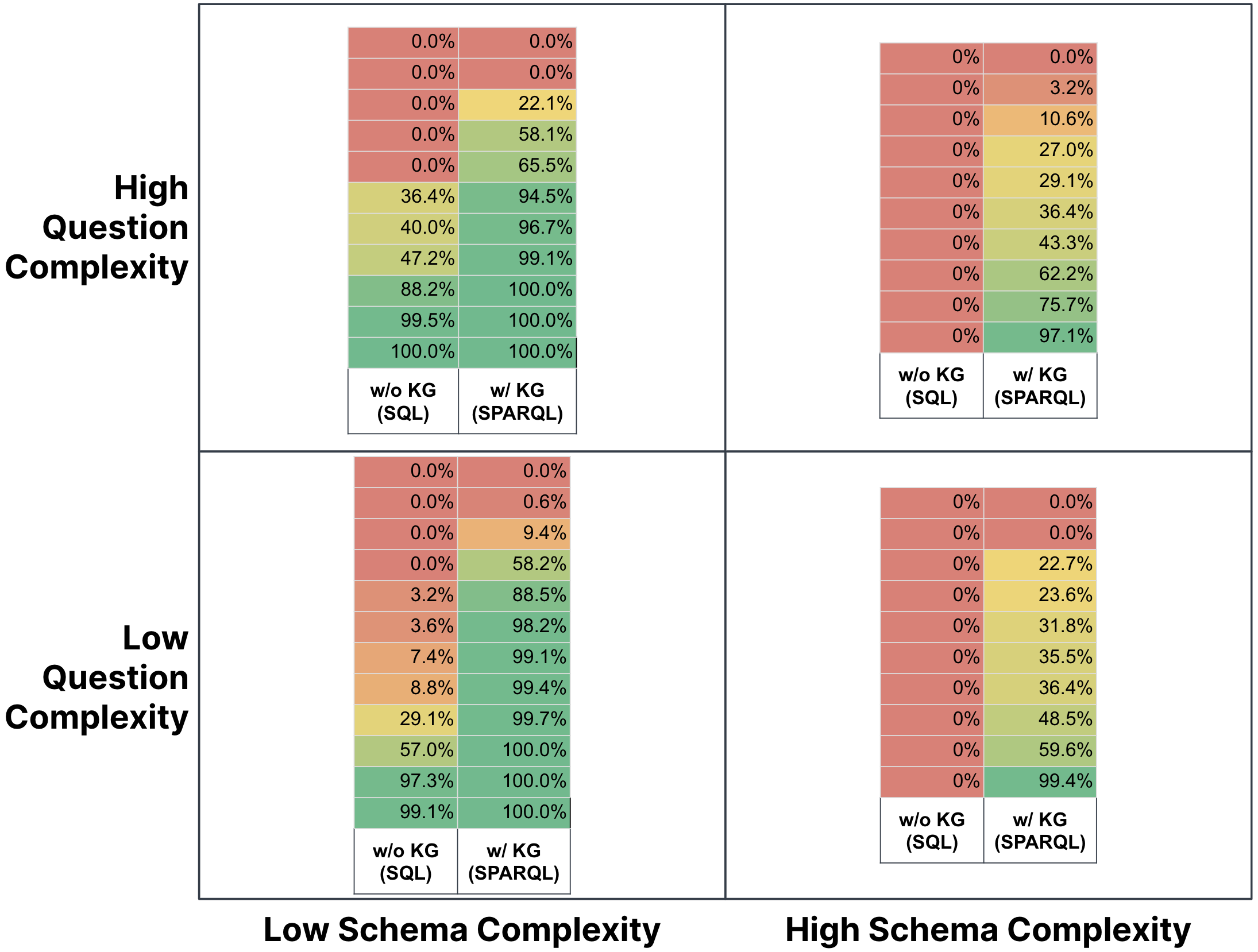}
\caption{Overall Execution Accuracy (OEA) of SPARQL and SQL for each quadrant as a heatmap}
\label{fig:fig4}
\end{figure}

Figure \ref{fig:fig4} presents the heat map for each quadrant.

Per the hypothesis, SPARQL achieves higher accuracy than SQL in every quadrant. 
Furthermore, it is surprising to observe that SQL was not able to answer any question in the High Schema Complexity quadrants. 

\paragraph{Discussion}
These results by quadrant sheds further light on understanding the extent. 
In each quadrant, SPARQL accuracy is higher than the SQL accuracy. 
While the SPARQL accuracy is 2.8X the SQL accuracy for Low Question/Low Schema and 1.8X for High Question/Low Schema, it was unforeseen that SQL was not able to accurately answer any questions for Low Question/High Schema and High Question/High Schema. 

The results also lead us to understand when SQL starts to fail.
When a question requires more than 5 tables to provide then answer, the accuracy drops to zero.

\subsection{Partial Accuracy}

To further investigate inaccurate answers, we manually analyzed the generated SQL and SPARQL queries. 
We observe that a subset of queries produced partially accurate results.
We consider a partially accurate answer to be one where the returned answers are accurate but incomplete. 

During the manual analysis, the following patterns for partially accurate answers are observed: 

\begin{itemize}
    \item \textbf{Overlap:} the columns returned by the query are correct, however, they are  a subset of the accurate answer. In some cases, they include other columns that are not part of the expected answer. This can be seen as a form of a semantic overlap\cite{DBLP:conf/dexa/GodfreyG99}. 
    \item \textbf{Return Identifier:} An internal identifier was returned instead of the appropriate label.   
\end{itemize}

Consider the question \textit{Return all the claims we have by claim number, open date and close date?} the following generated SQL and SPARQL query: 

\textbf{SQL}
\begin{verbatim}
SELECT Claim_Identifier, Claim_Open_Date, Claim_Close_Date
FROM Claim
\end{verbatim}

\textbf{SPARQL}
\begin{verbatim}
SELECT ?claim ?claimOpenDate ?claimCloseDate
WHERE {
    ?claim a in:Claim ;
           in:claimNumber ?claimNumber ;
           in:claimOpenDate ?claimOpenDate ;
           in:claimCloseDate ?claimCloseDate .
}
\end{verbatim}

The answer for claim open date and claim close date are accurate and is a subset of the correct answer. 
However in the SQL query, the \texttt{Claim\_Identifier} column is being returned as the claim number, when in fact, the claim number is actually the column \texttt{company\_claim\_number}. 
In the SPARQL query case, the variable \texttt{?claim} is returned which binds to the IRI that uniquely identifies each claim. 
The claim number is not returned.

Another issue we observed is that for a question that involved determining the average days between two dates, the generated SQL and SPARQL were both  semantically correct but the reason the query did not execute was due to a syntax error on date diff.

\paragraph{Discussion}
From a practical perspective, we believe it is important to have a measure of partial accuracy because it connects the results to reality. 
In practice, if a user is interacting with a system and the results are missing a column, they could ask for the missing column or provide a label instead of an identifier. 
Therefore partial accuracy may be acceptable for users. 
However this is an open question on how to define partial accuracy and how to score it.
Finally, this version of the benchmark did not include functions or operations over dates (see Section \ref{sec:ResearchAgenda} for a discussion on next steps on this topic).

\subsection{Inaccuracy}

During the manual analysis of the generated queries, we also observed query characteristics that generated the inaccurate answers. 
These characteristics were different for SQL and SPARQL. 

\subsubsection{SQL Inaccuracy}
The following three types of inaccuracies were observed: 
\begin{itemize}
    \item \textbf{Column Name Hallucinations}: Column names were generated that do not exist in the corresponding table. 
    \item \textbf{Value Hallucinations}: Generated value applied as a filter on a column where that value does not exist in the database.
    \item \textbf{Join Hallucinations}: Generated joins that are not accurate. 
\end{itemize}

\subsubsection{SPARQL Inaccuracy}

\begin{itemize}
    \item \textbf{Incorrect Path:} The generated query does not follow the correct path of the properties in the ontology. The generated path goes from A to C when the correct path is A to B to C. 
    \item \textbf{Incorrect Direction:} The generated query swaps the direction of a property. The generated direction is B to A, when the correct direction is A to B. 
\end{itemize}


\paragraph{Discussion}
The inaccuracy of SQL queries are based on hallucination while the inaccuracy of SPARQL queries are based on path inconsistency. 

The SQL hallucinations are evident: column names that don’t exist in a table and values that the LLM does not know if they exist. 
On the wrong joins, one can argue that the joins have been hallucinated because at an initial glance, the joins may seem plausible, but they are not how the database was designed, thus they have been made up by the LLM. 

For SPARQL queries, one can also argue that the paths were hallucinated because they are incorrect. 
However, the generated paths were indicative tht the LLM knew what the correct starting and end node was and the miss was on defining the correct path from the start node to the end node. 
One could even argue that the LLM appears to do some sort of reasoning but not always getting it correct.

To our surprise, while manually analyzing the generated SPARQL queries, we did not observe a hallucinated class or property. 
Anecdotally, we have observed that when a user goes to ChatGPT and asks a similar prompt, it does hallucinate properties in SPARQL query.
It is unclear why it did not hallucinate in the benchmark results. 
A speculation is that the prompt specifically stated \textit{return just the query, so it can be run verbatim from your response.} and this avoided hallucinations. 
However, the prompt structure is the same for SQL and SPARQL so the question remains, why did it hallucinate in SQL and not in SPARQL. 
See Section \ref{sec:ResearchAgenda} for further discussion on next steps. 

\section{Takeaways}

The results of the benchmark are evidence that support the hypothesis: \textit{An LLM powered question answering system that answers a Natural Language question over a Knowledge Graph representation of the SQL database returns more accurate results than a LLM powered question answering system that answers a Natural Language question over the SQL database without a knowledge graph.}

Given the results, we can provide an answer to our research question:

\textbf{RQ1:} To what extent Large Language Models (LLMs) can accurately answer Enterprise Natural Language questions over Enterprise SQL databases. 

\textbf{Answer:} Using GPT-4 and zero-shot prompting, Enterprise Natural Language questions over Enterprise SQL databases achieved 16.7\% Average Overall Execution Accuracy. For Low Question/Low Schema, the Overall Execution Accuracy was 25.5\%. For High Question/Low Schema, the Overall Execution Accuracy was 37.\%. However, for both Low Question/High Schema and High Question/High Schema, the accuracy was 0\%.

\textbf{RQ2:} To what extent Knowledge Graphs can improve the accuracy of Large Language Models (LLMs) to answer Enterprise Natural Language questions over Enterprise SQL databases. 

\textbf{Answer:} Using GPT-4 and zero-shot prompting, Enterprise Natural Language question over a Knowledge Graph representation of the enterprise SQL database achieved 54.2\% Average Overall Execution Accuracy. 
The overall SPARQL accuracy was 3x the SQL accuracy and the accuracy improvement was 37.5\%. 
For Low Question/Low Schema, the Overall Execution Accuracy was 71.1\%, which is 2.8X the SQL accuracy and 45.6\% accuracy improvement. 
For High Question/Low Schema, the Overall Execution Accuracy was 66.9\%, which is 1.8X the SQL accuracy and 29.5\% accuracy improvement. 
For Low Question/High Schema, the Overall Execution Accuracy was 35.7\% and the improvement was also 35.7\%. 
For High Question/High Schema, the Overall Execution Accuracy was 38.7\% and the improvement was also 38.7\%.

One can argue that this is a predictable result given that the SPARQL query over a Knowledge Graph representation of the SQL database contains all the required context. 
However, recall that our research question is to understand \textit{to what extent} Large Language Models (LLMs) can accurately answer Enterprise Natural Language questions over Enterprise SQL databases and how much that extent increases by using Knowledge Graphs. 
To the best of our knowledge, this extent was not understood until now. 

What is evident is that context is crucial for accuracy and these results further emphasizes the need to invest in context. 
The point to be made here is a call to action that investing in context of SQL databases is required to increase the accuracy of LLMs for question answering over the SQL databases. 

In the benchmark, the context was presented in the form of an ontology that describes the semantics of the business domain and mappings that connect the physical schema with the ontology which are used to create the knowledge graph. 
The ontology can also include further semantics such as synonms, labels in different languages, which are not expressible in a SQL DDL. 
We argue that this context needs to be treated as a first class citizen, thus, at minimum managed in a metadata management system (i.e. data catalog) and ideally on a knowledge graph architecture. 
Otherwise, the crucial context that provides accuracy would be managed in an ad-hoc manner. 



\section{Research Agenda}
\label{sec:ResearchAgenda}
Generative AI and Large Language Models are following the traditional hype cycle\footnote{\url{https://en.wikipedia.org/wiki/Gartner_hype_cycle}}. 
It is rapidly reaching the peak of inflated expectations.
We posit that applying these technologies to enterprise data will follow with the trough of disillusionment due to the lack of accuracy, explainability and governance. 
We, as many others, believe that the combination of Knowledge Graphs with Large Language Models will be a way to address the enterprise's concerns in order to hit the Slope of Enlightenment and ultimately the Plateau of Productivity. 

To separate the hype and noise from the substance and facts, we must conduct thorough research.
We believe that this benchmark is an important step to provide the facts that enterprises can use in order to navigate the hype and noise. 
The research agenda first comprises forthcoming steps to enhance the benchmark, followed by recommendations for future research topics. 
This research agenda is a call to action to unite the data management, knowledge graph and LLM communities.
The goal is to have continuous feedback and improvement of the benchmark. 
We encourage the community to suggest more addition of questions or techniques for evaluating LLMs.

\subsection{Benchmark Enhancements}

\paragraph{Benchmark Framework}
Each part of the benchmark framework can be enhanced. 

Enterprise SQL Schema:
\begin{itemize}
    \item This version is using a subset of the entire OMG P\&C Data Model for Property And Casualty Insurance, namely 13 tables. One could argue that this was not enough for the benchmark, however, we argue that the results are already representative with the current subset of the schema. Increasing the usage of the schema will shed further light on how much harder this problem is. 
    \item The instantiation of the database was done manually and consists of a couple of rows of data per table. Different instantiations of a database would identify scenarios of false positives. 
    \item The schema in this version is used to store and manage transactional data in third-normal form. How would the results change if the schema was modeled in different ways such as star schema, snowflake schema, data vault, one big table, Entity-Attribute-Value, etc? What happens if there are cycles in the schema? 
    \item The SQL DDL defines primary key/foreign key constraints. How much was that leveraged by the LLM? What happens if there are no constraints defined? What if more constraints are defined such as check constraint? 
\end{itemize}

Enterprise Questions
\begin{itemize}
    \item The amount of questions will never be enough. The current questions are projections and joins and lack filtering (i.e. during a period of time, in a specific value range, etc). 
    \item There can be various rewordings of the same natural language text that corresponds conceptually to the same question. 
    \item There can be questions that are ambiguous that can lead to different correct answers, depending on how the question is interpreted. Furthermore, questions may need to be followed up with further questions in order to disambiguate or suggest plausible interpretations if the initial question is ambiguous.
    \item Questions could not be answerable either because they are ambiguous or because of lack of data to answer it.  
    \item The high complex questions could be even more complex including inferencing, path and connectivity questions, etc.
    \item Questions can be provided with plausible but incorrect answers to assess the reliability of systems. 
    \item Questions can also be multi-turn conversations, where the follow up question uses the answer of a previous question.
    \item Questions can be in multiple languages.     
\end{itemize}

It is important to acknowledge that there can be biases in the benchmark questions.
We believe that addressing the aforementioned points, is a way to address that bias. 

Context Layer\begin{itemize}
    \item The current mappings consist of 1-1 table/column mappings, therefore they can be extended with different mapping patterns. Nevertheless, with the current mappings, a large inaccuracy gap has been identified. Possible mappings, based on \cite{DBLP:series/synthesis/2021Sequeda}: 
    \begin{itemize}
        \item Conditions: A subset of the rows in a table maps to a concept.
        \item Data as a Concept: A subset of the rows in a table, defined by an enumeration of values over a specific column, maps to a concept.
        \item Join Concept: A concept is mapped to multiple tables.
        \item Distinct: A table represents more than one concept and for a given concept, the rows that represent the concepts appear duplicated in the table. This appears when a table is denormalized.
        \item Concat: Multiple columns need to be concatenated in order to map to an attribute.
        \item Math: Values in different columns need to be applied in a math formula in order to be mapped to an attribute.
        \item Case: Values in a column are codes and each code needs to be mapped to a certain constant.
        \item Null: A column can have a NULL, which should be replaced by a constant in order to be mapped to an attribute.
        \item Join Attribute: The identifier of the concept is in one table while the column that needs to be mapped is in a different table.
    \end{itemize}
    \item The context is provided in RDF. It can also be provided in other graph formalisms such as OpenCypher and Gremlin. 
    \item Context can come from many different places including previous queries. Or even conversations from Slack/Teams? 
    \item The ontology can be more expressive to include hierarchies, cardinality constraints, enumeration, etc. For example, One Claim can have zero to one to many Claim Amounts, or A Claim Amount must be one of: Loss Payment, Expense Payment, Expense Reserve or Loss Reserve.
\end{itemize}

Scoring: 
\begin{itemize}
    \item The scoring could be extended to measure the semantic overlap between the answers in order to quantify the partial accuracy. 
    \item Measure time taken to produce an answer. This would be relevant in scenarios where chain-of-thought prompting approaches are used that require several calls to the LLMs to produce the final answer.
    \item The benchmark should follow the approach of Holistic Evaluation of Language Models (HELM)\footnote{\url{https://crfm.stanford.edu/helm/}}.
\end{itemize}

\paragraph{Benchmark Setup and Analysis} The prompting and models of the benchmark can be enhanced:

Prompting 
\begin{itemize}
    \item Current prompt is a simple zero-shot prompt. Different types of prompting strategies should be tested and compared such as few-shot, chain of thought, etc.
    \item The current size of the schema/ontology enables to put everything in a single context window. In practice, this may not feasible because the schema/ontology can be larger than the size of the context window\footnote{With GPT 4 Turbo, this now may be fully feasible}. And if it is feasible, what are the implications of accuracy with respect to a larger context? Thus we expect to have a RAG (Retrieval Augmented Generation) approach to extract the parts of the schema/ontology that are needed to then be provided into the prompt. 
\end{itemize}

Model
\begin{itemize}
    \item GPT-4 was the only model used in the experiments. Further testing should be done on a variety of open source (e.g. Llama) and closed models, including fine tuning. 
    \item SQL and Knowledge Graph foundation models should be evaluated such as NSQL\footnote{https://github.com/NumbersStationAI/NSQL} and SQLCoder\footnote{https://github.com/defog-ai/sqlcoder} for SQL and Ultra\footnote{https://github.com/DeepGraphLearning/ULTRA}\cite{galkin2023ultra} for Knowledge Graphs.  
\end{itemize}

\paragraph{Benchmark Analysis}
\begin{itemize}
    \item In addition to measuring the execution accuracy, the partial accuracy and inaccuracy should also be measured. In addition to defining a score of partial accuracy, the types of inaccuracies should be measured. For example, which type of hallucination appears the most?
    \item What correlations exist between Accuracy, Partial Accuracy and Inaccuracy and various query characteristics in SQL and SPARQL (amount of joins, length of paths, etc)
\end{itemize}

\subsection{Research Topics}

\paragraph{Business Implications}
What are the broader implications of the findings for the enterprise and their users?
For example, what would \textit{accurate enough} mean? 
Enterprise expect perfect results, but how do they know that their current approaches are accurate enough? 
These LLM-powered question answering systems show be compared to existing (manual) approaches that answer questions (i.e. the steps to answer a question by reviewing a dashboard). 
Do they achieve the same accuracy? What is the trust factor for each approach? How much time is spent to answer the question?

\paragraph{Qualitative Evaluation and User studies}
Given that humans are the ones interacting with these systems, we should study human interaction through user evaluation to go beyond accuracy. 
A user study can qualitatively evaluate users' satisfaction with answers and usability of the question answering system. 
For example, what are the implications of partial accurate answers? 
How does the size of the overlap impact user satisfaction?
Furthermore, there are various types of users that can be interacting with the system who have different context of the enterprise and intentions.
For example, a data analyst and an executive can be asking different types of questions. 
Additionally, the explanations should be tailored to the type of users.

\paragraph{Knowledge Engineering Methodologies}
A finding of our work is the the need to investing in context of SQL databases in the form of ontology and mappings which define the knowledge graph. 
Existing ontology and knowledge engineering methdologies have been defined in an era of smaller amount of data, and are manual processes\cite{DBLP:series/synthesis/2021Sequeda}. 
How can these methodologies be adapted to an era of large amounts of data and even incorporating LLMs in order increase productivity? 
Give that the call to action is to invest in knowledge graphs, how can we reduce the cost of that investment?

\paragraph{Diverse Domains}
Even though the benchmark is in the insurance domain, the framework can be applied to any other domain. 
Naturally, the schema, questions and context need to be defined by subject matter experts in a given domain. 
This can cover a broader spectrum of business sectors and help test LLMs accross different sectors. 
For specific industries, perhaps industry specific LLMs could be used such as BloombergGPT for the finance industry\cite{wu2023bloomberggpt}. 
Furthermore, this is an opportunity for enterprises who are implementing their own Question Answering system to apply the framework on their internal data and test the accuracy of their system using the benchmark.

\paragraph{Explainability}
Accuracy is the focus of this benchmark, which we consider to be the first critical step for LLMs to be reliable in the enterprise. 
Explainability is a next critical step such that the answers can be trusted by users. 
Therefore, we would like to extend the benchmark to understand the explainability capabilities of LLMs with and without Knowledge Graphs.  
We believe the following hypothesis is noteworthy: \textit{An LLM and Knowledge Graph powered question answering system provides explainability with higher user satisfaction compared to an LLM powered question answering system that doesn’t use knowledge graphs.}
The benchmark would need to be extended to include different types of explanations for different types of personas.
Additionally, the experiments would need to consist of user studies.

\paragraph{Cost}
The monetary cost is an aspect that should be measured in order to set the expectation on the return on investment for the enterprise. 
What are the monetary cost implications of achieving accurate answers? 
Can we find approaches that optimizes the cost? 
Or is the cost already relatively low that is not worth pursuing monetary cost optimization.

\paragraph{Pre and Post Processing}
In the current setup, the system attempts to immediately answer a question. 
However, what pre and post processing steps could be implemented that could improve accuracy? 
Could these be applied equivalently to both SQL and SPARQL? 
For example, in the pre-processing step, could it be determined if the data exist to answer the question? 
In a post processing step, after a query is generated, could it be possible to detect if the query will generate inaccurate results?

\section{Summary and Conclusion}
With the rise of Generative AI and Large Language Models (LLMs), question answering systems that implement Text-to-SQL has gained tremendous popularity. 
Furthermore, Knowledge Graphs has been acknowledge as having the potential for enhancing the accuracy of LLMs by providing missing business context and semantics.
The goal of this research is to understand the role of knowledge graphs on LLM's accuracy fo answering enterprise questiosn. 

This paper investigate two research questions: 1) to what extent Large Language Models (LLMs) can accurately answer enterprise natural language questions over enterprise SQL databases and 2) to what extent Knowledge Graphs can improve the accuracy of Large Language Models (LLMs) to answer
enterprise natural language questions over enterprise SQL databases. The hypothesis is that an LLM powered question answering system
that answers a Natural Language question over a Knowledge Graph representation of the SQL database returns more
accurate results than a LLM powered question answering system that answers a Natural Language question over the
SQL database without a knowledge graph.

In order to find answers to these research questions, we present a benchmark that takes in account enterprise SQL schema in the insurance domain, a series of enterprise questions that span a specturm of complexiy on questions and schema, and includes a context layer consiting of a business ontology and mappings which are used to build the knowledge graph. 

The results of benchmark determined that using GPT-4 and zero-shot prompting, Enterprise Natural Language questions over Enterprise SQL databases achieved 16.7\% while Enterprise Natural Language question over a Knowledge Graph representation of the enterprise SQL database achieved 54.2\% accuracy. 
Furthermore, for questions on low schema complexity, the accuracy using knowledge graphs was between 66.9\% and 71.1\%
while the accuracy without using knowledge graphs was between 25.5\% and 37.4\%.
Finally, questions on high schema complexity without using knowledge graph were not able to achieve any accurate responses, which the accuracy on knowledge graphs was between 35-38\%.
The results are evidence that support the hypothesis. 

\paragraph{Conclusion:} These experimental results are evidence that supports the main conclusion of this research: investing in Knowledge Graph provides higher accuracy for LLM powered question answering systems.
Therefore, if enterprises want to achieve higher accurate result in an LLM powered question answering system, they must treat the business context and semantics as a first class citizen and invest in a data catalog platform with a knowledge graph architecture. 

{\small
\paragraph{Acknowledgement}
We thank all our colleagues at data.world who supported our work. 
We are also extremley thankful for the early feedback we have received on this work from colleagues across industry and academia in the fields of AI, Databases and Knowledge Graphs: 
Albert Merono Penuela, 
Dan Bennett, 
Deborah McGuinness, 
Diego Collarana, 
Elena Simperl, 
Ethan Mollick, 
Gary George, 
Gary Marcus, 
George Fletcher, 
Luke Slotwinksi, 
Malcolm Chisholm, 
Mark Kitson, 
Michael Murray, 
Mike Dillinger, 
Mohammed Aaser,
Olaf Hartig, 
Omar Khawaja, 
Ora Lassila, 
Oscar Corcho, 
Patrick van de Belt, 
Paul Groth, 
Peter Lawerence, 
Rachel Wood, 
Steve Gustafan, 
Tony Seale, 
Vip Parmar.
}









\bibliographystyle{acm}






\newpage
\section{Appendix}

\subsection{Enteprise SQL Schema}
\label{Appendix:Schema}

\begin{verbatim}
CREATE TABLE Claim
( 
	Claim_Identifier     int  NOT NULL ,
	Catastrophe_Identifier int  NULL ,
	Claim_Description    varchar(5000)  NULL ,
	Claims_Made_Date     datetime  NULL ,
	Company_Claim_Number varchar(20)  NULL ,
	Company_Subclaim_Number varchar(5)  NULL ,
	Insurable_Object_Identifier int  NULL ,
	Occurrence_Identifier int  NULL ,
	Entry_Into_Claims_Made_Program_Date datetime  NULL ,
	Claim_Open_Date      datetime  NULL ,
	Claim_Close_Date     datetime  NULL ,
	Claim_Reopen_Date    datetime  NULL ,
	Claim_Status_Code    varchar(5)  NULL ,
	Claim_Reported_Date  datetime  NULL ,
	 PRIMARY KEY (Claim_Identifier ASC),
	 FOREIGN KEY (Catastrophe_Identifier) REFERENCES Catastrophe(Catastrophe_Identifier),
 FOREIGN KEY (Claim_Identifier) REFERENCES Claim(Claim_Identifier),
 FOREIGN KEY (Insurable_Object_Identifier) REFERENCES Insurable_Object(Insurable_Object_Identifier),
 FOREIGN KEY (Occurrence_Identifier) REFERENCES Occurrence(Occurrence_Identifier)
)

CREATE TABLE Claim_Amount
( 
	Claim_Amount_Identifier bigint  NOT NULL ,
	Claim_Identifier     int  NOT NULL ,
	Claim_Offer_Identifier int  NULL ,
	Amount_Type_Code     varchar(20)  NULL ,
	Event_Date           datetime  NULL ,
	Claim_Amount         decimal(15,2)  NULL ,
	Insurance_Type_Code  char(1)  NULL ,
	 PRIMARY KEY (Claim_Amount_Identifier ASC),
	 FOREIGN KEY (Claim_Offer_Identifier) REFERENCES Claim_Offer(Claim_Offer_Identifier),
 FOREIGN KEY (Claim_Identifier) REFERENCES Claim(Claim_Identifier)
)

CREATE TABLE Loss_Payment
( 
	Claim_Amount_Identifier bigint  NOT NULL ,
	 PRIMARY KEY (Claim_Amount_Identifier ASC),
	 FOREIGN KEY (Claim_Amount_Identifier) REFERENCES Claim_Payment(Claim_Amount_Identifier)
)

CREATE TABLE Loss_Reserve
( 
	Claim_Amount_Identifier bigint  NOT NULL ,
	 PRIMARY KEY (Claim_Amount_Identifier ASC),
	 FOREIGN KEY (Claim_Amount_Identifier) REFERENCES Claim_Reserve(Claim_Amount_Identifier)
)

CREATE TABLE Expense_Payment
( 
	Claim_Amount_Identifier bigint  NOT NULL ,
	 PRIMARY KEY (Claim_Amount_Identifier ASC),
	 FOREIGN KEY (Claim_Amount_Identifier) REFERENCES Claim_Payment(Claim_Amount_Identifier)
)

CREATE TABLE Expense_Reserve
( 
	Claim_Amount_Identifier bigint  NOT NULL ,
	 PRIMARY KEY (Claim_Amount_Identifier ASC),
	 FOREIGN KEY (Claim_Amount_Identifier) REFERENCES Claim_Reserve(Claim_Amount_Identifier)
)

CREATE TABLE Claim_Coverage
( 
	Claim_Identifier     int  NOT NULL ,
	Effective_Date       datetime  NOT NULL ,
	Policy_Coverage_Detail_Identifier int  NOT NULL ,
	 PRIMARY KEY (Claim_Identifier ASC,Effective_Date ASC,Policy_Coverage_Detail_Identifier ASC),
	 FOREIGN KEY (Claim_Identifier) REFERENCES Claim(Claim_Identifier),
 FOREIGN KEY (Effective_Date,Policy_Coverage_Detail_Identifier) 
   REFERENCES Policy_Coverage_Detail(Effective_Date,Policy_Coverage_Detail_Identifier)
)

CREATE TABLE Policy_Coverage_Detail
( 
	Effective_Date       datetime  NOT NULL ,
	Policy_Coverage_Detail_Identifier int  NOT NULL ,
	Coverage_Identifier  int  NOT NULL ,
	Insurable_Object_Identifier int  NOT NULL ,
	Policy_Identifier    int  NOT NULL ,
	Coverage_Part_Code   varchar(20)  NOT NULL ,
	Coverage_Description varchar(2000)  NULL ,
	Expiration_Date      datetime  NULL ,
	Coverage_Inclusion_Exclusion_Code char(1)  NULL ,
	 PRIMARY KEY (Effective_Date ASC,Policy_Coverage_Detail_Identifier ASC),
	 FOREIGN KEY (Insurable_Object_Identifier) REFERENCES Insurable_Object(Insurable_Object_Identifier),
 FOREIGN KEY (Coverage_Identifier) REFERENCES Coverage(Coverage_Identifier),
 FOREIGN KEY (Coverage_Part_Code,Policy_Identifier) 
   REFERENCES Policy_Coverage_Part(Coverage_Part_Code,Policy_Identifier)
)

CREATE TABLE Policy
( 
	Policy_Identifier    int  NOT NULL ,
	Effective_Date       datetime  NULL ,
	Expiration_Date      datetime  NULL ,
	Policy_Number        varchar(50)  NULL ,
	Status_Code          varchar(20)  NULL ,
	Geographic_Location_Identifier int  NULL ,
	 PRIMARY KEY (Policy_Identifier ASC),
	 FOREIGN KEY (Geographic_Location_Identifier) 
    REFERENCES Geographic_Location(Geographic_Location_Identifier),
 FOREIGN KEY (Policy_Identifier) REFERENCES Agreement(Agreement_Identifier)
)

CREATE TABLE Policy_Amount
( 
	Policy_Amount_Identifier bigint  NOT NULL ,
	Geographic_Location_Identifier int  NOT NULL ,
	Policy_Identifier    int  NULL ,
	Effective_Date       datetime  NULL ,
	Amount_Type_Code     varchar(5)  NULL ,
	Earning_Begin_Date   datetime  NULL ,
	Earning_End_Date     datetime  NULL ,
	Policy_Coverage_Detail_Identifier int  NULL ,
	Policy_Amount        decimal(15,2)  NULL ,
	Insurable_Object_Identifier int  NULL ,
	Insurance_Type_Code  char(1)  NULL ,
	 PRIMARY KEY (Policy_Amount_Identifier ASC),
	 FOREIGN KEY (Effective_Date,Policy_Coverage_Detail_Identifier) 
  REFERENCES Policy_Coverage_Detail(Effective_Date,Policy_Coverage_Detail_Identifier),
 FOREIGN KEY (Policy_Identifier) 
   REFERENCES Policy(Policy_Identifier),
 FOREIGN KEY (Geographic_Location_Identifier) 
   REFERENCES Geographic_Location(Geographic_Location_Identifier),
 FOREIGN KEY (Insurable_Object_Identifier) 
   REFERENCES Insurable_Object(Insurable_Object_Identifier)
)

CREATE TABLE Agreement_Party_Role
( 
	Agreement_Identifier int  NOT NULL ,
	Party_Identifier     bigint  NOT NULL ,
	Party_Role_Code      varchar(20)  NOT NULL ,
	Effective_Date       datetime  NOT NULL ,
	Expiration_Date      datetime  NULL ,
	 PRIMARY KEY (Agreement_Identifier ASC,Party_Identifier ASC,Party_Role_Code ASC,Effective_Date ASC),
	 FOREIGN KEY (Agreement_Identifier) REFERENCES Agreement(Agreement_Identifier),
 FOREIGN KEY (Party_Identifier) REFERENCES Party(Party_Identifier),
 FOREIGN KEY (Party_Role_Code) REFERENCES Party_Role(Party_Role_Code)
)

CREATE TABLE Premium
( 
	Policy_Amount_Identifier bigint  NOT NULL ,
	 PRIMARY KEY (Policy_Amount_Identifier ASC),
	 FOREIGN KEY (Policy_Amount_Identifier) REFERENCES Policy_Amount(Policy_Amount_Identifier)
)

CREATE TABLE Catastrophe
( 
	Catastrophe_Identifier int  NOT NULL ,
	Catastrophe_Type_Code varchar(20)  NULL ,
	Catastrophe_Name     varchar(100)  NULL ,
	Industry_Catastrophe_Code varchar(20)  NULL ,
	Company_Catastrophe_Code varchar(20)  NULL ,
	 PRIMARY KEY (Catastrophe_Identifier ASC)
)

\end{verbatim}

\subsection{Enteprise Questions}
\label{Appendix:Questions}

\textbf{Low Question/Low Schema Complexity} 
\begin{itemize}
\item What are all the premiums that have been paid by policy holders?
\item Return all the policies and their policy holder by id
\item Return all the claims we have by claim number, open date and close date?
\item What is the premium amount of all policies by policy number?
\item Return all the policies and the agents that sold them by policy number and agent id
\item What is the premium amount of all policies by policy number, coverage effective date and coverage expiration date
\item What are the loss payment amount of all claims by claim number?
\item What are the loss reserve amount of all claims by claim number?
\item What are the expense reserve amount of all claims by claim number?
\item What are the expense payment amount of all claims by claim number?
\item What are all our policies that have a claim associated to them by policy and claim number?
\item Return all the policies we have by policy number, effective date and expiration date?
\end{itemize}

\textbf{High Question/Low Schema Complexity}
\begin{itemize}
\item What is the total amount of premiums that a policy holder has paid by policy number?
\item What is the average time to settle a claim by policy number?
\item What is the total amount of premiums that a policy holder has paid?
\item How many policies have agents sold by agent id?
\item What is the total loss amounts, which is the sum of loss payment, loss reserve amount by claim number?
\item How many policies does each policy holder have by policy holder id?
\item What is the total amount of premiums paid by policy number?
\item How many claims have been placed by policy number?
\item What is the average policy size which is the the total amount of premium divided by the number of policies?
\item How many policies do we have?
\item How many claims do we have?
\end{itemize}

\textbf{Low Question/High Schema Complexity}
\begin{itemize}
\item Return policy holders and the claims they have made and the correspoinding catastrophe
\item Return agents and the policy they have sold that have had a claim and the corresponding catastrophe it had.
\item Return agents and the policy they have sold that have had a claim and the corresponding loss reserve amount by agent id, policy number and claim number
\item What are the loss payment, loss reserve, expense payment, expense reserve amount by claim number and corresponding policy number, policy holder, premium amount paid, the catastrophe it had, and the agent who sold it?
\item What are the loss payment, loss reserve, expense payment, expense reserve amount by claim number and corresponding policy number, policy holder and premium amount paid?
\item What are the loss payment, loss reserve, expense payment, expense reserve amount by claim number and corresponding policy number, policy holder, premium amount paid and the agent who sold it?
\item Return agents and the policies they have sold that have had a claim and the corresponding loss payment amount by agent id, policy number and claim number
\item Return agents and the policy they have sold that have had a claim and the corresponding expense reserve amount by agent id, policy number and claim number
\item Return agents and the policy they have sold that have had a claim and the corresponding expense payment amount by agent id, policy number and claim number
\item What are the loss payment, loss reserve, expense payment, expense reserve amount by claim number?
\end{itemize}

\textbf{High Question/High Schema Complexity}
\begin{itemize}
\item What is the loss ratio of each policy and agent who sold it by policy number and agent id?
\item What are the total loss, which is the sum of loss payment, loss reserve, expense payment, expense reserve amount by claim number and corresponding policy number, policy holder and premium amount paid?
\item What are the total loss, which is the sum of loss payment, loss reserve, expense payment, expense reserve amount by claim number, catastrophe and corresponding policy number?
\item What is the loss ratio, number of claims, total loss by policy number and premium where total loss is the sum of loss payment, loss reserve, expense payment, expense reserve amount and loss ratio is total loss divided by premium?
\item What is the total loss of each claim by claim number where total loss is the sum of loss payment, loss reserve, expense payment, expense reserve amount?
\item What is the total loss of each policy that an agent has sold by agent id where total loss is the sum of loss payment, loss reserve, expense payment, expense reserve amounts?
\item What is the average loss of each policy by policy number and number of claims where loss is the sum of loss payment, loss reserve, expense payment, expense reserve amounts?
\item What is the total loss of each policy by policy number where total loss is the sum of loss payment, loss reserve, expense payment, expense reserve amounts?
\item What is the average loss of each policy by policy number where loss is the sum of loss payment, loss reserve, expense payment, expense reserve amounts?
\item What are the total loss, which is the sum of loss payment, loss reserve, expense payment, expense reserve amount by catastrophe and policy number?
\end{itemize}

\subsection{Context Layer}
\label{Appendix:Context}

\subsubsection{Ontology in OWL}
\begin{verbatim}

@prefix : <http://data.world/schema/insurance/> .
@prefix dc: <http://purl.org/dc/elements/1.1/> .
@prefix in: <http://data.world/schema/insurance/> .
@prefix owl: <http://www.w3.org/2002/07/owl#> .
@prefix rdf: <http://www.w3.org/1999/02/22-rdf-syntax-ns#> .
@prefix xml: <http://www.w3.org/XML/1998/namespace> .
@prefix xsd: <http://www.w3.org/2001/XMLSchema#> .
@prefix foaf: <http://xmlns.com/foaf/spec/> .
@prefix rdfs: <http://www.w3.org/2000/01/rdf-schema#> .
@prefix skos: <http://www.w3.org/2004/02/skos/core#> .
@base <http://data.world/schema/insurance/> .

<http://data.world/schema/insurance/> rdf:type owl:Ontology .

#################################################################
#    Object Properties
#################################################################

###  http://data.world/schema/insurance/against
in:against rdf:type owl:ObjectProperty ;
           rdfs:domain in:Claim ;
           rdfs:range in:PolicyCoverageDetail ;
           rdfs:isDefinedBy <http://data.world/schema/insurance/> ;
           rdfs:label "against" .


###  http://data.world/schema/insurance/hasCatastrophe
in:hasCatastrophe rdf:type owl:ObjectProperty ;
                  rdfs:domain in:Claim ;
                  rdfs:range in:Catastrophe ;
                  rdfs:isDefinedBy <http://data.world/schema/insurance/> ;
                  rdfs:label "has catastrophe" .


###  http://data.world/schema/insurance/hasExpensePayment
in:hasExpensePayment rdf:type owl:ObjectProperty ;
                     rdfs:domain in:Claim ;
                     rdfs:range in:ExpensePayment ;
                     rdfs:isDefinedBy <http://data.world/schema/insurance/> ;
                     rdfs:label "has expense payment" .


###  http://data.world/schema/insurance/hasExpenseReserve
in:hasExpenseReserve rdf:type owl:ObjectProperty ;
                     rdfs:domain in:Claim ;
                     rdfs:range in:ExpenseReserve ;
                     rdfs:isDefinedBy <http://data.world/schema/insurance/> ;
                     rdfs:label "has expense reserve" .


###  http://data.world/schema/insurance/hasLossPayment
in:hasLossPayment rdf:type owl:ObjectProperty ;
                  rdfs:domain in:Claim ;
                  rdfs:range in:LossPayment ;
                  rdfs:isDefinedBy <http://data.world/schema/insurance/> ;
                  rdfs:label "has loss payment" .


###  http://data.world/schema/insurance/hasLossReserve
in:hasLossReserve rdf:type owl:ObjectProperty ;
                  rdfs:domain in:Claim ;
                  rdfs:range in:LossReserve ;
                  rdfs:isDefinedBy <http://data.world/schema/insurance/> ;
                  rdfs:label "has loss reserve" .


###  http://data.world/schema/insurance/hasPolicy
in:hasPolicy rdf:type owl:ObjectProperty ;
             rdfs:domain in:PolicyCoverageDetail ;
             rdfs:range in:Policy ;
             rdfs:isDefinedBy <http://data.world/schema/insurance/> ;
             rdfs:label "has policy" .


###  http://data.world/schema/insurance/hasPolicyHolder
in:hasPolicyHolder rdf:type owl:ObjectProperty ;
                   rdfs:domain in:Policy ;
                   rdfs:range in:PolicyHolder ;
                   rdfs:isDefinedBy <http://data.world/schema/insurance/> ;
                   rdfs:label "has policy holder" .


###  http://data.world/schema/insurance/hasPremiumAmount
in:hasPremiumAmount rdf:type owl:ObjectProperty ;
                    rdfs:domain in:PolicyCoverageDetail ;
                    rdfs:range in:Premium ;
                    rdfs:isDefinedBy <http://data.world/schema/insurance/> ;
                    rdfs:label "has premium amount" .


###  http://data.world/schema/insurance/soldByAgent
in:soldByAgent rdf:type owl:ObjectProperty ;
               rdfs:domain in:Policy ;
               rdfs:range in:Agent ;
               rdfs:isDefinedBy <http://data.world/schema/insurance/> ;
               rdfs:label "sold by agent" .


#################################################################
#    Data properties
#################################################################

###  http://data.world/schema/insurance/agentId
in:agentId rdf:type owl:DatatypeProperty ;
           rdfs:domain in:Agent ;
           rdfs:isDefinedBy <http://data.world/schema/insurance/> ;
           rdfs:label "Agent ID" .


###  http://data.world/schema/insurance/catastropheName
in:catastropheName rdf:type owl:DatatypeProperty ;
                   rdfs:domain in:Catastrophe ;
                   rdfs:isDefinedBy <http://data.world/schema/insurance/> ;
                   rdfs:label "Catastrophe Name" .


###  http://data.world/schema/insurance/claimCloseDate
in:claimCloseDate rdf:type owl:DatatypeProperty ;
                  rdfs:domain in:Claim ;
                  rdfs:range xsd:dateTime ;
                  rdfs:isDefinedBy <http://data.world/schema/insurance/> ;
                  rdfs:label "Claim Close Date" .


###  http://data.world/schema/insurance/claimNumber
in:claimNumber rdf:type owl:DatatypeProperty ;
               rdfs:domain in:Claim ;
               rdfs:isDefinedBy <http://data.world/schema/insurance/> ;
               rdfs:label "Claim Number" .


###  http://data.world/schema/insurance/claimOpenDate
in:claimOpenDate rdf:type owl:DatatypeProperty ;
                 rdfs:domain in:Claim ;
                 rdfs:range xsd:dateTime ;
                 rdfs:isDefinedBy <http://data.world/schema/insurance/> ;
                 rdfs:label "Claim Open Date" .


###  http://data.world/schema/insurance/expensePaymentAmount
in:expensePaymentAmount rdf:type owl:DatatypeProperty ;
                        rdfs:domain in:ExpensePayment ;
                        rdfs:isDefinedBy <http://data.world/schema/insurance/> ;
                        rdfs:label "Expense Payment Amount" .


###  http://data.world/schema/insurance/expenseReserveAmount
in:expenseReserveAmount rdf:type owl:DatatypeProperty ;
                        rdfs:domain in:ExpenseReserve ;
                        rdfs:isDefinedBy <http://data.world/schema/insurance/> ;
                        rdfs:label "Expense Reserve Amount" .


###  http://data.world/schema/insurance/lossPaymentAmount
in:lossPaymentAmount rdf:type owl:DatatypeProperty ;
                     rdfs:domain in:LossPayment ;
                     rdfs:isDefinedBy <http://data.world/schema/insurance/> ;
                     rdfs:label "Loss Payment Amount" .


###  http://data.world/schema/insurance/lossReserveAmount
in:lossReserveAmount rdf:type owl:DatatypeProperty ;
                     rdfs:domain in:LossReserve ;
                     rdfs:isDefinedBy <http://data.world/schema/insurance/> ;
                     rdfs:label "Loss Reserve Amount" .


###  http://data.world/schema/insurance/policyCoverageEffectiveDate
in:policyCoverageEffectiveDate rdf:type owl:DatatypeProperty ;
                               rdfs:domain in:PolicyCoverageDetail ;
                               rdfs:range xsd:dateTime ;
                               rdfs:isDefinedBy <http://data.world/schema/insurance/> ;
                               rdfs:label "Policy Coverage Effective Date" .


###  http://data.world/schema/insurance/policyCoverageExpirationDate
in:policyCoverageExpirationDate rdf:type owl:DatatypeProperty ;
                                rdfs:domain in:PolicyCoverageDetail ;
                                rdfs:range xsd:dateTime ;
                                rdfs:isDefinedBy <http://data.world/schema/insurance/> ;
                                rdfs:label "Policy Coverage Expiration Date" .


###  http://data.world/schema/insurance/policyEffectiveDate
in:policyEffectiveDate rdf:type owl:DatatypeProperty ;
                       rdfs:domain in:Policy ;
                       rdfs:range xsd:dateTime ;
                       rdfs:isDefinedBy <http://data.world/schema/insurance/> ;
                       rdfs:label "Policy Effective Date" .


###  http://data.world/schema/insurance/policyExpirationDate
in:policyExpirationDate rdf:type owl:DatatypeProperty ;
                        rdfs:domain in:Policy ;
                        rdfs:range xsd:dateTime ;
                        rdfs:isDefinedBy <http://data.world/schema/insurance/> ;
                        rdfs:label "Policy Expiration Date" .


###  http://data.world/schema/insurance/policyHolderId
in:policyHolderId rdf:type owl:DatatypeProperty ;
                  rdfs:domain in:PolicyHolder ;
                  rdfs:isDefinedBy <http://data.world/schema/insurance/> ;
                  rdfs:label "Policy Holder ID" .


###  http://data.world/schema/insurance/policyNumber
in:policyNumber rdf:type owl:DatatypeProperty ;
                rdfs:domain in:Policy ;
                rdfs:isDefinedBy <http://data.world/schema/insurance/> ;
                rdfs:label "Policy Number" .


###  http://data.world/schema/insurance/premiumAmount
in:premiumAmount rdf:type owl:DatatypeProperty ;
                 rdfs:domain in:Premium ;
                 rdfs:isDefinedBy <http://data.world/schema/insurance/> ;
                 rdfs:label "Premium Amount" .


###  http://data.world/schema/insurance/premiumAmountMonthly
in:premiumAmountMonthly rdf:type owl:DatatypeProperty ;
                        rdfs:domain in:Premium ;
                        rdfs:isDefinedBy <http://data.world/schema/insurance/> ;
                        rdfs:label "Premium Amount Monthly" .


###  http://data.world/schema/insurance/premiumPeriod
in:premiumPeriod rdf:type owl:DatatypeProperty ;
                 rdfs:domain in:Premium ;
                 rdfs:isDefinedBy <http://data.world/schema/insurance/> ;
                 rdfs:label "Premium Period" .


#################################################################
#    Classes
#################################################################

###  http://data.world/schema/insurance/Agent
in:Agent rdf:type owl:Class ;
         rdfs:isDefinedBy <http://data.world/schema/insurance/> ;
         rdfs:label "Agent" .


###  http://data.world/schema/insurance/Catastrophe
in:Catastrophe rdf:type owl:Class ;
               rdfs:isDefinedBy <http://data.world/schema/insurance/> ;
               rdfs:label "Catastrophe" .


###  http://data.world/schema/insurance/Claim
in:Claim rdf:type owl:Class ;
         rdfs:isDefinedBy <http://data.world/schema/insurance/> ;
         rdfs:label "Claim" .


###  http://data.world/schema/insurance/ExpensePayment
in:ExpensePayment rdf:type owl:Class ;
                  rdfs:isDefinedBy <http://data.world/schema/insurance/> ;
                  rdfs:label "Expense Payment" .


###  http://data.world/schema/insurance/ExpenseReserve
in:ExpenseReserve rdf:type owl:Class ;
                  rdfs:isDefinedBy <http://data.world/schema/insurance/> ;
                  rdfs:label "Expense Reserve" .


###  http://data.world/schema/insurance/LossPayment
in:LossPayment rdf:type owl:Class ;
               rdfs:isDefinedBy <http://data.world/schema/insurance/> ;
               rdfs:label "Loss Payment" .


###  http://data.world/schema/insurance/LossReserve
in:LossReserve rdf:type owl:Class ;
               rdfs:isDefinedBy <http://data.world/schema/insurance/> ;
               rdfs:label "Loss Reserve" .


###  http://data.world/schema/insurance/Policy
in:Policy rdf:type owl:Class ;
          rdfs:isDefinedBy <http://data.world/schema/insurance/> ;
          rdfs:label "Policy" .


###  http://data.world/schema/insurance/PolicyCoverageDetail
in:PolicyCoverageDetail rdf:type owl:Class ;
                        rdfs:isDefinedBy <http://data.world/schema/insurance/> ;
                        rdfs:label "Policy Coverage Detail" .


###  http://data.world/schema/insurance/PolicyHolder
in:PolicyHolder rdf:type owl:Class ;
                rdfs:isDefinedBy <http://data.world/schema/insurance/> ;
                rdfs:label "Policy Holder" .


###  http://data.world/schema/insurance/Premium
in:Premium rdf:type owl:Class ;
           rdfs:isDefinedBy <http://data.world/schema/insurance/> ;
           rdfs:label "Premium" .


#################################################################
#    Annotations
#################################################################

<http://data.world/schema/insurance/> rdfs:comment "This ontology defines Business Concepts, 
Attributes, and Relationships in the insurance domain. It is inspired by OMG Property and Casualty 
Data Model." ;
                                      rdfs:label "Insurance Ontology" .
    
\end{verbatim}

\subsubsection{Mapping in R2RML}

\begin{tiny}
\begin{verbatim}
@prefix rr:  <http://www.w3.org/ns/r2rml#> .
@prefix map:  <http://capsenta.com/mappings#> .
@prefix dwo:  <https://ontology.data.world/v0#> .


### Claim
map:TripleMap_Claim_16 a rr:TriplesMap ;
    rr:subjectMap          [ rr:class <http://data.world/schema/insurance/Claim> ;
                           rr:template "https://myinsurancecompany.linked.data.world/d/omg-pc-database/Claim-{Claim_Identifier}" ] ;
    rr:logicalTable         [ rr:tableName "myinsurancecompany.omg-pc-database.claim" ] .

map:TripleMap_ClaimNumber_1 a rr:TriplesMap ;
    rr:predicateObjectMap  [ rr:objectMap [ rr:column "company_claim_number" ] ;
                           rr:predicate <http://data.world/schema/insurance/claimNumber> ] ;
    rr:subjectMap          [ rr:template "https://myinsurancecompany.linked.data.world/d/omg-pc-database/Claim-{Claim_Identifier}" ] ;
    rr:logicalTable         [ rr:tableName "myinsurancecompany.omg-pc-database.claim" ] .

map:TripleMap_ClaimCloseDate_8 a rr:TriplesMap ;
    rr:predicateObjectMap  [ rr:objectMap [ rr:column "claim_close_date" ] ;
                           rr:predicate <http://data.world/schema/insurance/claimCloseDate> ] ;
    rr:subjectMap          [ rr:template "https://myinsurancecompany.linked.data.world/d/omg-pc-database/Claim-{Claim_Identifier}" ] ;
    rr:logicalTable         [ rr:tableName "myinsurancecompany.omg-pc-database.claim" ] .

map:TripleMap_ClaimOpenDate_10 a rr:TriplesMap ;
    rr:predicateObjectMap  [ rr:objectMap [ rr:column "claim_open_date" ] ;
                           rr:predicate <http://data.world/schema/insurance/claimOpenDate> ] ;
    rr:subjectMap          [ rr:template "https://myinsurancecompany.linked.data.world/d/omg-pc-database/Claim-{Claim_Identifier}" ] ;
    rr:logicalTable         [ rr:tableName "myinsurancecompany.omg-pc-database.claim" ] .    

### Policy
map:TripleMap_Policy_13 a rr:TriplesMap ;
    rr:subjectMap          [ rr:class <http://data.world/schema/insurance/Policy> ;
                           rr:template "https://myinsurancecompany.linked.data.world/d/omg-pc-database/Policy-{policy_identifier}" ] ;
    rr:logicalTable         [ rr:tableName "myinsurancecompany.omg-pc-database.policy" ] .

map:TripleMap_PolicyExpirationDate_15 a rr:TriplesMap ;
    rr:predicateObjectMap  [ rr:objectMap [ rr:column "expiration_date" ] ;
                           rr:predicate <http://data.world/schema/insurance/policyExpirationDate> ] ;
    rr:subjectMap          [ rr:template "https://myinsurancecompany.linked.data.world/d/omg-pc-database/Policy-{policy_identifier}" ] ;
    rr:logicalTable         [ rr:tableName "myinsurancecompany.omg-pc-database.policy" ] .

map:TripleMap_PolicyNumber_11 a rr:TriplesMap ;
    rr:predicateObjectMap  [ rr:objectMap [ rr:column "policy_number" ] ;
                           rr:predicate <http://data.world/schema/insurance/policyNumber> ] ;
    rr:subjectMap          [ rr:template "https://myinsurancecompany.linked.data.world/d/omg-pc-database/Policy-{policy_identifier}" ] ;
    rr:logicalTable         [ rr:tableName "myinsurancecompany.omg-pc-database.policy" ] .

map:TripleMap_PolicyEffectiveDate_15 a rr:TriplesMap ;
    rr:predicateObjectMap  [ rr:objectMap [ rr:column "effective_date" ] ;
                           rr:predicate <http://data.world/schema/insurance/policyEffectiveDate> ] ;
    rr:subjectMap          [ rr:template "https://myinsurancecompany.linked.data.world/d/omg-pc-database/Policy-{policy_identifier}" ] ;
    rr:logicalTable         [ rr:tableName "myinsurancecompany.omg-pc-database.policy" ] .

### PolicyCoverageDetail
map:TripleMap_PolicyCoverageDetail_17 a rr:TriplesMap ;
    rr:subjectMap          [ rr:class <http://data.world/schema/insurance/PolicyCoverageDetail> ;
                           rr:template "https://myinsurancecompany.linked.data.world/d/omg-pc-database/PolicyCoverageDetail-{policy_coverage_detail_identifier}" ] ;
    rr:logicalTable         [ rr:tableName "myinsurancecompany.omg-pc-database.policy_coverage_detail" ] .


map:TripleMap_PolicyCoverageExpirationDate_0 a rr:TriplesMap ;
    rr:predicateObjectMap  [ rr:objectMap [ rr:column "expiration_date" ] ;
                           rr:predicate <http://data.world/schema/insurance/policyCoverageExpirationDate> ] ;
    rr:subjectMap          [ rr:template "https://myinsurancecompany.linked.data.world/d/omg-pc-database/PolicyCoverageDetail-{policy_coverage_detail_identifier}" ] ;
    rr:logicalTable         [ rr:tableName "myinsurancecompany.omg-pc-database.policy_coverage_detail" ] .


map:TripleMap_PolicyCoverageEffectiveDate_19 a rr:TriplesMap ;
    rr:predicateObjectMap  [ rr:objectMap [ rr:column "effective_date" ] ;
                           rr:predicate <http://data.world/schema/insurance/policyCoverageEffectiveDate> ] ;
    rr:subjectMap          [ rr:template "https://myinsurancecompany.linked.data.world/d/omg-pc-database/PolicyCoverageDetail-{policy_coverage_detail_identifier}" ] ;
    rr:logicalTable         [ rr:tableName "myinsurancecompany.omg-pc-database.policy_coverage_detail" ] .


map:TripleMap_PolicyCoverageDetailID_3 a rr:TriplesMap ;
    rr:predicateObjectMap  [ rr:objectMap [ rr:column "policy_coverage_detail_identifier" ] ;
                           rr:predicate <http://data.world/schema/insurance/policyCoverageDetailId> ] ;
    rr:subjectMap          [ rr:template "https://myinsurancecompany.linked.data.world/d/omg-pc-database/PolicyCoverageDetail-{policy_coverage_detail_identifier}" ] ;
    rr:logicalTable         [ rr:tableName "myinsurancecompany.omg-pc-database.policy_coverage_detail" ] .


### Catastrophe
map:TripleMap_Catastrophe_0 a rr:TriplesMap ;
    rr:subjectMap          [ rr:class <http://data.world/schema/insurance/Catastrophe> ;
                           rr:template "https://myinsurancecompany.linked.data.world/d/omg-pc-database/Catastrophe-{catastrophe_identifier}" ] ;
    rr:logicalTable         [ rr:tableName "myinsurancecompany.omg-pc-database.catastrophe" ] .

map:TripleMap_CatastropheName_18 a rr:TriplesMap ;
    rr:predicateObjectMap  [ rr:objectMap [ rr:column "catastrophe_name" ] ;
                           rr:predicate <http://data.world/schema/insurance/catastropheName> ] ;
    rr:subjectMap          [ rr:template "https://myinsurancecompany.linked.data.world/d/omg-pc-database/Catastrophe-{catastrophe_identifier}" ] ;
    rr:logicalTable   [ rr:tableName "myinsurancecompany.omg-pc-database.catastrophe" ] .


### Agent
map:TripleMap_Agent_0 a rr:TriplesMap ;
    rr:subjectMap          [ rr:class <http://data.world/schema/insurance/Agent> ;
                           rr:template "https://myinsurancecompany.linked.data.world/d/omg-pc-database/Agent-{party_identifier}" ] ;
    rr:logicalTable         [ rr:sqlQuery """select distinct party_identifier 
from agreement_party_role
join policy on agreement_party_role.agreement_identifier = policy.policy_identifier 
where agreement_party_role.party_role_code = 'AG'""" ;
                              dwo:queryBase <https://myinsurancecompany.linked.data.world/d/omg-pc-database/> ] .

map:TripleMap_AgentID_2 a rr:TriplesMap ;
    rr:predicateObjectMap  [ rr:objectMap [ rr:column "party_identifier" ] ;
                           rr:predicate <http://data.world/schema/insurance/agentId> ] ;
    rr:subjectMap          [ rr:template "https://myinsurancecompany.linked.data.world/d/omg-pc-database/Agent-{party_identifier}" ] ;
    rr:logicalTable         [ rr:sqlQuery """select distinct party_identifier 
from agreement_party_role
join policy on agreement_party_role.agreement_identifier = policy.policy_identifier 
where agreement_party_role.party_role_code = 'AG'""" ;
                              dwo:queryBase <https://myinsurancecompany.linked.data.world/d/omg-pc-database/> ] .

### PolicyHolder

map:TripleMap_PolicyHolder_12 a rr:TriplesMap ;
    rr:subjectMap          [ rr:class <http://data.world/schema/insurance/PolicyHolder> ;
                           rr:template "https://myinsurancecompany.linked.data.world/d/omg-pc-database/Policy-Holder-{party_identifier}" ] ;
    rr:logicalTable         [ rr:sqlQuery """select distinct party_identifier 
from agreement_party_role
join policy on agreement_party_role.agreement_identifier = policy.policy_identifier 
where agreement_party_role.party_role_code = 'PH'""" ;
                              dwo:queryBase <https://myinsurancecompany.linked.data.world/d/omg-pc-database/> ] .


map:TripleMap_PolicyHolderID_12 a rr:TriplesMap ;
    rr:predicateObjectMap  [ rr:objectMap [ rr:column "party_identifier" ] ;
                           rr:predicate <http://data.world/schema/insurance/policyHolderId> ] ;
    rr:subjectMap          [ rr:template "https://myinsurancecompany.linked.data.world/d/omg-pc-database/Policy-Holder-{party_identifier}" ] ;
    rr:logicalTable         [ rr:sqlQuery """select distinct party_identifier 
from agreement_party_role
join policy on agreement_party_role.agreement_identifier = policy.policy_identifier 
where agreement_party_role.party_role_code = 'PH'""" ;
                              dwo:queryBase <https://myinsurancecompany.linked.data.world/d/omg-pc-database/> ] .

### Premium
map:TripleMap_Premium_6 a rr:TriplesMap ;
    rr:subjectMap          [ rr:class <http://data.world/schema/insurance/Premium> ;
                           rr:template "https://myinsurancecompany.linked.data.world/d/omg-pc-database/Premium-{policy_amount_identifier}" ] ;
    rr:logicalTable         [ rr:sqlQuery """select policy_amount.policy_amount_identifier, 
    policy_amount.policy_amount, 
    policy_amount.amount_type_code , 
    case when policy_amount.amount_type_code = 'Year' then policy_amount.policy_amount/12 end as monthly_policy_amount, 
    policy_amount.policy_coverage_detail_identifier
from policy_amount 
join premium on policy_amount.policy_amount_identifier = premium.policy_amount_identifier""" ;
                              dwo:queryBase <https://myinsurancecompany.linked.data.world/d/omg-pc-database/> ] .

map:TripleMap_PremiumAmount_14 a rr:TriplesMap ;
    rr:predicateObjectMap  [ rr:objectMap [ rr:column "policy_amount" ] ;
                           rr:predicate <http://data.world/schema/insurance/premiumAmount> ] ;
    rr:subjectMap          [ rr:template "https://myinsurancecompany.linked.data.world/d/omg-pc-database/Premium-{policy_amount_identifier}" ] ;
    rr:logicalTable         [ rr:sqlQuery """select policy_amount.policy_amount_identifier, 
    policy_amount.policy_amount, 
    policy_amount.amount_type_code , 
    case when policy_amount.amount_type_code = 'Year' then policy_amount.policy_amount/12 end as monthly_policy_amount, 
    policy_amount.policy_coverage_detail_identifier
from policy_amount 
join premium on policy_amount.policy_amount_identifier = premium.policy_amount_identifier""" ;
                              dwo:queryBase <https://myinsurancecompany.linked.data.world/d/omg-pc-database/> ] .


### Relationships
      
map:TripleMap_against_14 a rr:TriplesMap ;
    rr:predicateObjectMap  [ rr:objectMap [ rr:template "https://myinsurancecompany.linked.data.world/d/omg-pc-database/PolicyCoverageDetail-{policy_coverage_detail_identifier}" ] ;
                           rr:predicate <http://data.world/schema/insurance/against> ] ;
    rr:subjectMap          [ rr:template "https://myinsurancecompany.linked.data.world/d/omg-pc-database/Claim-{claim_identifier}" ] ;
    rr:logicalTable         [ rr:tableName "myinsurancecompany.omg-pc-database.claim_coverage" ] .      

map:TripleMap_haspolicycoveragedetail_15 a rr:TriplesMap ;
    rr:predicateObjectMap  [ rr:objectMap [ rr:template "https://myinsurancecompany.linked.data.world/d/omg-pc-database/PolicyCoverageDetail-{policy_coverage_detail_identifier}" ] ;
                           rr:predicate <http://data.world/schema/insurance/hasPolicyCoverageDetail> ] ;
    rr:subjectMap          [ rr:template "https://myinsurancecompany.linked.data.world/d/omg-pc-database/Policy-{policy_identifier}" ] ;
    rr:logicalTable         [ rr:tableName "myinsurancecompany.omg-pc-database.policy_coverage_detail" ] .


map:TripleMap_hasclaim_8 a rr:TriplesMap ;
    rr:predicateObjectMap  [ rr:objectMap [ rr:template "https://myinsurancecompany.linked.data.world/d/omg-pc-database/Claim-{claim_identifier}" ] ;
                           rr:predicate <http://data.world/schema/insurance/hasClaim> ] ;
    rr:subjectMap          [ rr:template "https://myinsurancecompany.linked.data.world/d/omg-pc-database/PolicyCoverageDetail-{policy_coverage_detail_identifier}" ] ;
    rr:logicalTable         [ rr:tableName "myinsurancecompany.omg-pc-database.claim_coverage" ] .

map:TripleMap_hascatastrophe_19 a rr:TriplesMap ;
    rr:predicateObjectMap  [ rr:objectMap [ rr:template "https://myinsurancecompany.linked.data.world/d/omg-pc-database/Catastrophe-{catastrophe_identifier}" ] ;
                           rr:predicate <http://data.world/schema/insurance/hasCatastrophe> ] ;
    rr:subjectMap          [ rr:template "https://myinsurancecompany.linked.data.world/d/omg-pc-database/Claim-{claim_identifier}" ] ;
    rr:logicalTable         [ rr:tableName "myinsurancecompany.omg-pc-database.claim" ] .


map:TripleMap_haspremium_0 a rr:TriplesMap ;
    rr:predicateObjectMap  [ rr:objectMap [ rr:template "https://myinsurancecompany.linked.data.world/d/omg-pc-database/Premium-{policy_amount_identifier}" ] ;
                           rr:predicate <http://data.world/schema/insurance/hasPremiumAmount> ] ;
    rr:subjectMap          [ rr:template "https://myinsurancecompany.linked.data.world/d/omg-pc-database/PolicyCoverageDetail-{policy_coverage_detail_identifier}" ] ;
    rr:logicalTable         [ rr:sqlQuery """select policy_amount.policy_amount_identifier, 
    policy_amount.policy_amount, 
    policy_amount.amount_type_code , 
    case when policy_amount.amount_type_code = 'Year' then policy_amount.policy_amount/12 end as monthly_policy_amount, 
    policy_amount.policy_coverage_detail_identifier
from policy_amount 
join premium on policy_amount.policy_amount_identifier = premium.policy_amount_identifier""" ;
                              dwo:queryBase <https://myinsurancecompany.linked.data.world/d/omg-pc-database/> ] .

map:TripleMap_haspolicyholder_18 a rr:TriplesMap ;
    rr:predicateObjectMap  [ rr:objectMap [ rr:template "https://myinsurancecompany.linked.data.world/d/omg-pc-database/PolicyHolder-{party_identifier}" ] ;
                           rr:predicate <http://data.world/schema/insurance/hasPolicyHolder> ] ;
    rr:subjectMap          [ rr:template "https://myinsurancecompany.linked.data.world/d/omg-pc-database/Policy-{policy_identifier}" ] ;
    rr:logicalTable         [ rr:sqlQuery """select policy.policy_identifier, party_identifier 
from agreement_party_role
join policy on agreement_party_role.agreement_identifier = policy.policy_identifier
where agreement_party_role.party_role_code = 'PH'""" ;
                              dwo:queryBase <https://myinsurancecompany.linked.data.world/d/omg-pc-database/> ] .

map:TripleMap_haspolicy_1 a rr:TriplesMap ;
    rr:predicateObjectMap  [ rr:objectMap [ rr:template "https://myinsurancecompany.linked.data.world/d/omg-pc-database/Policy-{policy_identifier}" ] ;
                           rr:predicate <http://data.world/schema/insurance/hasPolicy> ] ;
    rr:subjectMap          [ rr:template "https://myinsurancecompany.linked.data.world/d/omg-pc-database/PolicyCoverageDetail-{policy_coverage_detail_identifier}" ] ;
    rr:logicalTable         [ rr:tableName "myinsurancecompany.omg-pc-database.policy_coverage_detail" ] .

map:TripleMap_soldbyagent_14 a rr:TriplesMap ;
    rr:predicateObjectMap  [ rr:objectMap [ rr:template "https://myinsurancecompany.linked.data.world/d/omg-pc-database/Agent-{party_identifier}" ] ;
                           rr:predicate <http://data.world/schema/insurance/soldByAgent> ] ;
    rr:subjectMap          [ rr:template "https://myinsurancecompany.linked.data.world/d/omg-pc-database/Policy-{policy_identifier}" ] ;
    rr:logicalTable         [ rr:sqlQuery """select policy.policy_identifier, party_identifier 
from agreement_party_role
join policy on agreement_party_role.agreement_identifier = policy.policy_identifier
where agreement_party_role.party_role_code = 'AG'""" ;
                              dwo:queryBase <https://myinsurancecompany.linked.data.world/d/omg-pc-database/> ] .


map:TripleMap_hasexpensepayment_2 a rr:TriplesMap ;
    rr:predicateObjectMap  [ rr:objectMap [ rr:template "https://myinsurancecompany.linked.data.world/d/omg-pc-database/ClaimAmount-{claim_amount_identifier}" ] ;
                           rr:predicate <http://data.world/schema/insurance/hasExpensePayment> ] ;
    rr:subjectMap          [ rr:template "https://myinsurancecompany.linked.data.world/d/omg-pc-database/Claim-{claim_identifier}" ] ;
    rr:logicalTable         [ rr:sqlQuery """select ca.claim_identifier, ca.claim_amount_identifier, ca.claim_amount
from claim_amount ca
join expense_payment ep on ca.claim_amount_identifier = ep.claim_amount_identifier""" ;
                              dwo:queryBase <https://myinsurancecompany.linked.data.world/d/omg-pc-database/> ] .

map:TripleMap_ExpensePaymentAmount_7 a rr:TriplesMap ;
    rr:predicateObjectMap  [ rr:objectMap [ rr:column "claim_amount" ] ;
                           rr:predicate <http://data.world/schema/insurance/expensePaymentAmount> ] ;
    rr:subjectMap          [ rr:template "https://myinsurancecompany.linked.data.world/d/omg-pc-database/ClaimAmount-{claim_amount_identifier}" ] ;
    rr:logicalTable         [ rr:sqlQuery """select ca.claim_identifier, ca.claim_amount_identifier, ca.claim_amount
from claim_amount ca
join expense_payment ep on ca.claim_amount_identifier = ep.claim_amount_identifier""" ;
                              dwo:queryBase <https://myinsurancecompany.linked.data.world/d/omg-pc-database/> ] .

map:TripleMap_haslossreserve_0 a rr:TriplesMap ;
    rr:predicateObjectMap  [ rr:objectMap [ rr:template "https://myinsurancecompany.linked.data.world/d/omg-pc-database/ClaimAmount-{claim_amount_identifier}" ] ;
                           rr:predicate <http://data.world/schema/insurance/hasLossReserve> ] ;
    rr:subjectMap          [ rr:template "https://myinsurancecompany.linked.data.world/d/omg-pc-database/Claim-{claim_identifier}" ] ;
    rr:logicalTable         [ rr:sqlQuery """select ca.claim_identifier, ca.claim_amount_identifier, ca.claim_amount
from claim_amount ca
join loss_reserve lr on ca.claim_amount_identifier = lr.claim_amount_identifier""" ;
                              dwo:queryBase <https://myinsurancecompany.linked.data.world/d/omg-pc-database/> ] .

map:TripleMap_LossReserveAmount_17 a rr:TriplesMap ;
    rr:predicateObjectMap  [ rr:objectMap [ rr:column "claim_amount" ] ;
                           rr:predicate <http://data.world/schema/insurance/lossReserveAmount> ] ;
    rr:subjectMap          [ rr:template "https://myinsurancecompany.linked.data.world/d/omg-pc-database/ClaimAmount-{claim_amount_identifier}" ] ;
    rr:logicalTable         [ rr:sqlQuery """select ca.claim_identifier, ca.claim_amount_identifier, ca.claim_amount
from claim_amount ca
join loss_reserve lr on ca.claim_amount_identifier = lr.claim_amount_identifier""" ;
                              dwo:queryBase <https://myinsurancecompany.linked.data.world/d/omg-pc-database/> ] .

map:TripleMap_hasexpensereserve_16 a rr:TriplesMap ;
    rr:predicateObjectMap  [ rr:objectMap [ rr:template "https://myinsurancecompany.linked.data.world/d/omg-pc-database/ClaimAmount-{claim_amount_identifier}" ] ;
                           rr:predicate <http://data.world/schema/insurance/hasExpenseReserve> ] ;
    rr:subjectMap          [ rr:template "https://myinsurancecompany.linked.data.world/d/omg-pc-database/Claim-{claim_identifier}" ] ;
    rr:logicalTable         [ rr:sqlQuery """select ca.claim_identifier, ca.claim_amount_identifier, ca.claim_amount
from claim_amount ca
join expense_reserve er on ca.claim_amount_identifier = er.claim_amount_identifier""" ;
                              dwo:queryBase <https://myinsurancecompany.linked.data.world/d/omg-pc-database/> ] .

map:TripleMap_ExpenseReserveAmount_9 a rr:TriplesMap ;
    rr:predicateObjectMap  [ rr:objectMap [ rr:column "claim_amount" ] ;
                           rr:predicate <http://data.world/schema/insurance/expenseReserveAmount> ] ;
    rr:subjectMap          [ rr:template "https://myinsurancecompany.linked.data.world/d/omg-pc-database/ClaimAmount-{claim_amount_identifier}" ] ;
    rr:logicalTable         [ rr:sqlQuery """select ca.claim_identifier, ca.claim_amount_identifier, ca.claim_amount
from claim_amount ca
join expense_reserve er on ca.claim_amount_identifier = er.claim_amount_identifier""" ;
                              dwo:queryBase <https://myinsurancecompany.linked.data.world/d/omg-pc-database/> ] .

map:TripleMap_haslosspayment_13 a rr:TriplesMap ;
    rr:predicateObjectMap  [ rr:objectMap [ rr:template "https://myinsurancecompany.linked.data.world/d/omg-pc-database/ClaimAmount-{claim_amount_identifier}" ] ;
                           rr:predicate <http://data.world/schema/insurance/hasLossPayment> ] ;
    rr:subjectMap          [ rr:template "https://myinsurancecompany.linked.data.world/d/omg-pc-database/Claim-{claim_identifier}" ] ;
    rr:logicalTable         [ rr:sqlQuery """select ca.claim_identifier, ca.claim_amount_identifier, ca.claim_amount
from claim_amount ca
join loss_payment lp on ca.claim_amount_identifier = lp.claim_amount_identifier""" ;
                              dwo:queryBase <https://myinsurancecompany.linked.data.world/d/omg-pc-database/> ] .


map:TripleMap_LossPaymentAmount_15 a rr:TriplesMap ;
    rr:predicateObjectMap  [ rr:objectMap [ rr:column "claim_amount" ] ;
                           rr:predicate <http://data.world/schema/insurance/lossPaymentAmount> ] ;
    rr:subjectMap          [ rr:template "https://myinsurancecompany.linked.data.world/d/omg-pc-database/ClaimAmount-{claim_amount_identifier}" ] ;
    rr:logicalTable         [ rr:sqlQuery """select ca.claim_identifier, ca.claim_amount_identifier, ca.claim_amount
from claim_amount ca
join loss_payment lp on ca.claim_amount_identifier = lp.claim_amount_identifier""" ;
                              dwo:queryBase <https://myinsurancecompany.linked.data.world/d/omg-pc-database/> ] .

\end{verbatim}
\end{tiny}

\end{document}